\documentclass[10pt,twocolumn,letterpaper]{article}
\usepackage[pagenumbers]{iccv} 
\definecolor{iccvblue}{rgb}{0.21,0.49,0.74}
\usepackage[pagebackref,breaklinks,colorlinks,allcolors=iccvblue]{hyperref}

\definecolor{darkgreen}{rgb}{0,0.5,0}
\definecolor{purple}{rgb}{0.5,0,0.5}
\definecolor{royalblue}{rgb}{0.25,0.4,1}
\definecolor{darkyellow}{rgb}{0.75,0.75,0}

\newcommand{\ours}{\textsf{\textbf{Strefer}} }
\newcommand{\ourseos}{\textsf{\textbf{Strefer}}}

\newcommand{\modeleos}{\textsf{\textbf{Strefer}}}

\usepackage[most]{tcolorbox}
\usepackage{amsmath}
\usepackage{algorithm}
\usepackage{algpseudocode}
\usepackage{graphicx}
\usepackage{colortbl}
\usepackage{amsmath}
\usepackage{multirow}
\usepackage{tablefootnote}
\usepackage{makecell}
\definecolor{LightSkyBlue}{RGB}{135,206,250}
\definecolor{lightblue}{RGB}{230, 230, 255}
\definecolor{LightGray}{RGB}{101, 101, 105}
\definecolor{LightPink}{RGB}{203, 66, 245}
\definecolor{highlight_g}{RGB}{60, 180, 75}  
\definecolor{highlight_y}{RGB}{255, 225, 25}  
\definecolor{highlight_b}{RGB}{0, 130, 200}  
\definecolor{highlight_o}{RGB}{245, 130, 49}  
\definecolor{highlight5}{RGB}{145, 30, 180}

\makeatletter
\renewcommand\@dotsep{10000}
\makeatother

\def\paperID{21} 
\def\confName{ICCV}
\def\confYear{2025}

\title{\modeleos: Empowering Video LLMs with \textit{S}pace-\textit{T}ime \textit{Refer}ring and Reasoning via Synthetic Instruction Data}

\author{
  \vspace{-20pt}\\
  Honglu Zhou, Xiangyu Peng, Shrikant Kendre, Michael S. Ryoo, Silvio Savarese,\\
  Caiming Xiong, Juan Carlos Niebles\\
  Salesforce AI Research\\
  \vspace{-10pt}\\
  \href{https://strefer.github.io/}{\Large{\texttt{https://strefer.github.io}}} \\
}

\begin{document}

\twocolumn[{
  \renewcommand\twocolumn[1][]{#1}
  \maketitle

  \begin{center}
    \centering
    \captionsetup{type=figure}
    \vspace{-22pt}
    \includegraphics[scale=0.8]{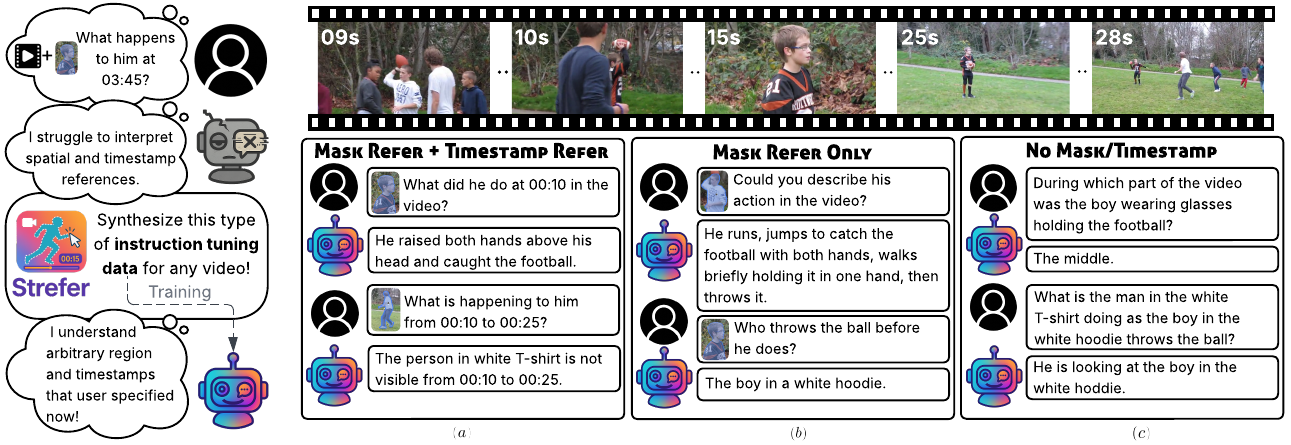}
    \vspace{-20pt}
    \captionof{figure}{Our goal is to synthesize instruction-response pairs through a scalable, grounded approach that enhances fine-grained spatial and temporal perception and reasoning over videos for tuning Video LLMs.
    We hypothesize that the video space-time referring task, which requires understanding user-specified regions within a video at particular moments or over defined time intervals, can greatly enhance 
    the effectiveness and versatility of 
    video analysis with Video LLMs, facilitating advanced applications in areas including navigation, surveillance, and interactive robotics. To this end, we introduce \ours (Figure~\ref{fig:method}), which 
    synthesizes instruction data aligned with our goals. \ours produces instruction-response pairs for: \textit{(a)} queries with mask and timestamp references, \textit{(b)} queries with mask (or timestamp) references only, and \textit{(c)} queries without either.   
Though current implementation of \ours does not any use proprietary models, without the need to annotate large volumes of new videos, instruction data from \ours empowers models for space-time referring and spatiotemporal reasoning (\textit{ref.}~Table~\ref{tab:res_region_cap}, ~\ref{tab:res_region_qa}, and~\ref{tab:res_more}). 
Using \ourseos, we generated $947,854$ instruction data points from just $4,253$ NExT-QA~\cite{xiao2021next} videos (test set excluded; up to 3 minutes long).
While
our \colorbox{LightSkyBlue!20}{final recipe},
consisted exclusively of short videos a few minutes long and added only $545$ more videos compared to the baseline,  
it led to performance gains across multiple benchmarks.
    }
    \label{fig:teaser}
  \end{center}
}]

\begin{abstract}

\vspace{-10pt}
Next-generation AI companions must go beyond general video understanding to resolve spatial and temporal references in dynamic, real-world environments.
Existing Video Large Language Models (Video LLMs), while capable of coarse-level comprehension, struggle with fine-grained, spatiotemporal reasoning, especially when user queries rely on time-based event references for temporal anchoring, or gestural cues for spatial anchoring to clarify object references and positions.
To bridge this 
critical
gap, we introduce \ourseos, a synthetic instruction data generation framework designed to equip Video LLMs with spatiotemporal referring and reasoning capabilities. \ours produces diverse instruction-tuning data using a data engine that pseudo-annotates temporally dense, fine-grained video metadata, capturing rich spatial and temporal information in a structured manner, including subjects, objects, their locations as masklets, and their action descriptions and timelines. Our approach enhances the ability of Video LLMs to interpret spatial and temporal references, fostering more versatile, space-time-aware reasoning essential for real-world AI companions. Without using proprietary models, costly human annotation, or the need to annotate large volumes of new videos, experimental evaluations show that models trained with data produced by \ours outperform baselines on tasks requiring spatial and temporal disambiguation.
Additionally, these models exhibit enhanced space-time-aware reasoning, establishing a new foundation for perceptually grounded, instruction-tuned Video LLMs.

\end{abstract}

\begin{figure*}[t]
	\centering
	\vspace{-60pt}
    \includegraphics[scale=0.73]{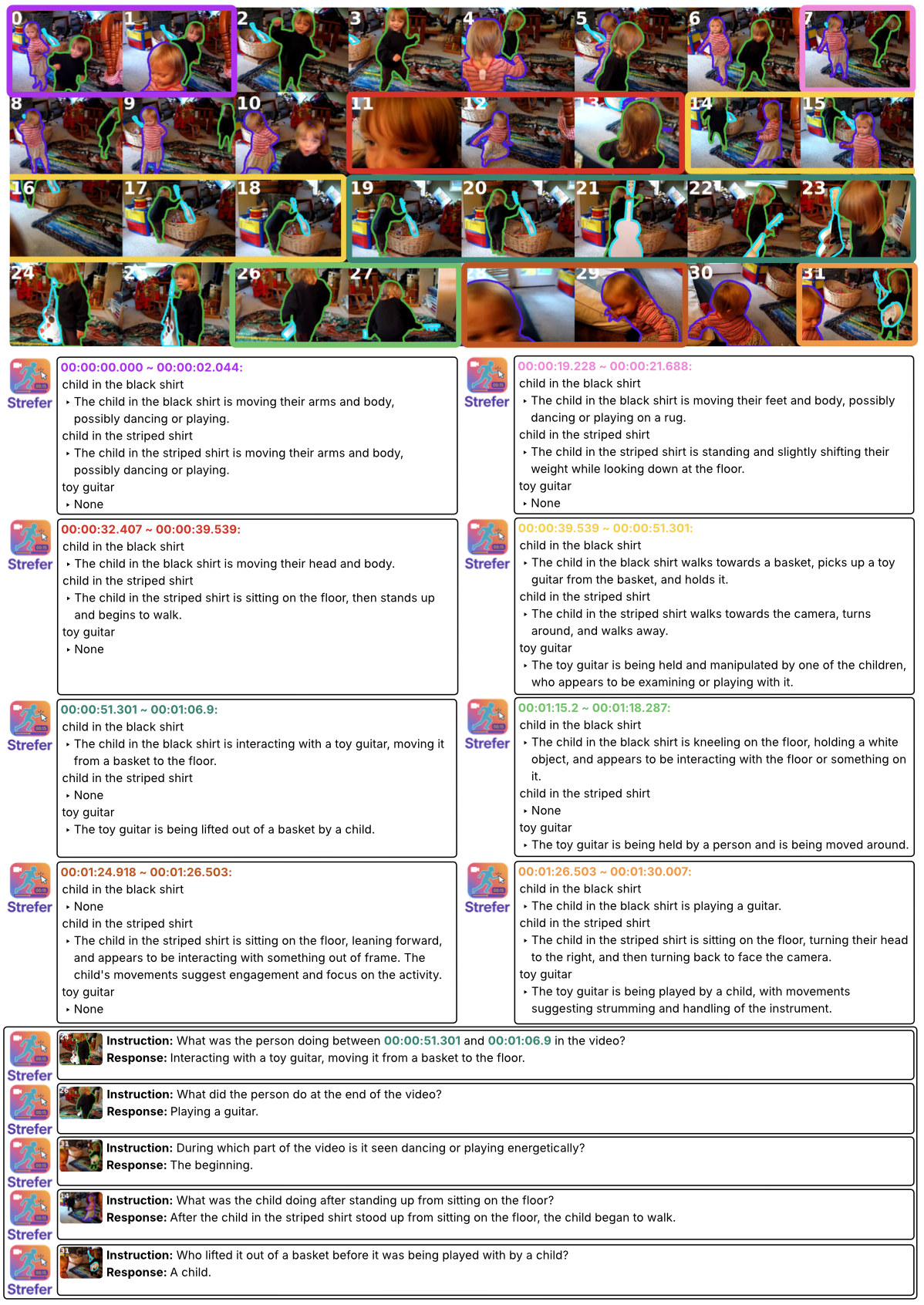}
    \vspace{-5pt}
\captionof{figure}{Example of \ourseos-annotated instruction-response pairs (bottom) and video metadata (top).
Each instruction begins with the prefix: \textit{Please answer the following question about the $<$region$>$} (omitted in the figure). For each instruction-response pair, 
the boundary of object mask referred to by \textit{$<$region$>$} is shown beside the pair.
\ours automatically clips the video into segments and pseudo-annotates the video metadata, including active entities, their locations (as masklets), and their action descriptions and timelines for complex video scenarios, such as scenes containing multiple entities of the same category, and cases where entities do not appear in the first frame, or temporarily exit and re-enter the frame.
Based on the auto-generated video metadata, \ours produces instruction-response pairs
requiring no legacy annotations or manual efforts. 
}
\vspace{-50pt}
    \label{fig:quali_good_ourdata_2}
\end{figure*}

\clearpage
\onecolumn                 

\begin{center}
    {\Large \bf Table of Contents}
\end{center}
\vspace{0.8cm}
{\hypersetup{linkcolor=black}\normalsize \tableofcontents}

\clearpage
\twocolumn                 
\pagenumbering{arabic}     
\setcounter{page}{4}

\section{Introduction}
\vspace{-5pt}

The vision of real-life AI assistants as everyday companions is rapidly becoming a reality.
To function seamlessly in real-world environments, these agents must understand human queries that are grounded in both space and time. 
People often rely on non-verbal cues such as gestures that
clarify which object is being referred to, or time-based references tied to specific moments in the past. For instance, a user might point at a cup and say, ``Can you bring me that cup?'', a request that may be ambiguous without detailed verbal clarification if several similar looking cups are nearby. They may also ask questions using  time-based references: ``At 11 a.m.~today, whom did I talk to at the market?'' 
These scenarios highlight the critical need for space-time referring capabilities, where an AI agent must resolve queries with references that are not purely linguistic, but situated in dynamic spatio-temporal contexts.

Recent Video Large Language Models (Video LLMs) have shown promise in general video understanding~\cite{zhang2025videollama,bai2025qwen2,chen2024expanding,wang2024tarsier,xu2024pllava,li2024llavaonevision,hong2024cogvlm2,chung2025unifying,xu2024slowfast,cheng2024videollama,li2024llamavid,blip3video}. 
However, these models tend to operate at a coarse level, lacking the granularity required to track and reason about specific object states, movements, and temporal relations~\cite{cho2025perceptionlm,liu2024tempcompass}. Their limitations are especially evident in videos with complex spatial and temporal structures, where multiple entities interact and change over time. They are also incapable of handling user queries that are not purely verbal but contain specific space-time references (\textit{ref.}~Fig.~\ref{fig:teaser}). This shortcoming stems not only from architectural constraints, but more significantly, from the scarcity of fine-grained, object-level instruction tuning data focused on 
spatial and temporal
understanding, referring and reasoning within complex videos.

\begin{figure*}[t]
	\centering
	\vspace{-10pt}
    \includegraphics[scale=0.85]{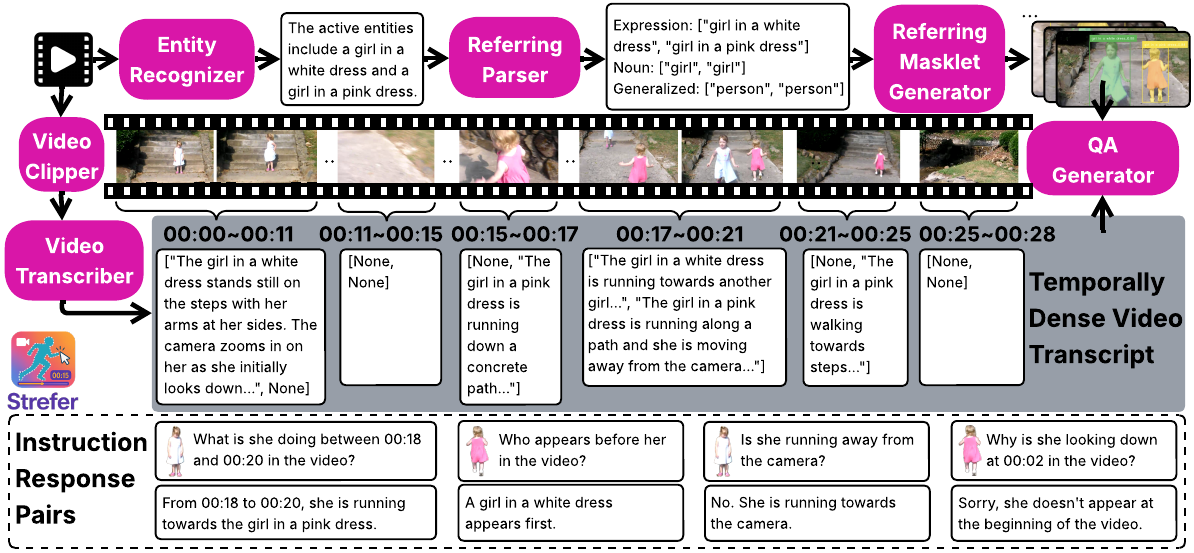}
	\caption{We introduce \ourseos, a novel data engine that automatically generates synthetic instruction data—without manual effort or legacy annotation—featuring multimodal prompts grounded in complex spatiotemporal video structures, designed to train Video LLMs for space-time referring and reasoning tasks (Sec.~\ref{subsec:auto_data_engine}).
    By design, \ours handles challenging scenarios—such as scenes containing multiple entities of the same category and cases where entities do not appear in the first frame or temporarily exit and re-enter the frame—while scaling to minutes-long videos beyond the scope of existing auto-data engines in the video domain.
    The current implementation of \ours does not use any proprietary models. The \textit{Entity Recognizer} and \textit{Video Transcriber} are Video LLMs (Tarsier-34b~\cite{wang2024tarsier}), the \textit{Referring Parser} is an LLM (Qwen2.5-32B-Instruct~\cite{qwen2.5}), and the \textit{QA Generator} uses either templates
    or an LLM (Qwen2.5-32B-Instruct~\cite{qwen2.5}). The \textit{Video Clipper} is based on PySceneDetect~\cite{scenedetect2025}, SigLIP~\cite{zhai2023sigmoid}, and hierarchical clustering; and Fig.~\ref{fig:r2m} illustrates our novel \textit{Referring Masklet Generator}, leveraging GroundingDINO~\cite{liu2023groundingdino}, SAM2~\cite{sam2}, and RexSeek~\cite{rexseek}.
    Without the need to annotate large volumes of new videos, instruction data from \ours empowers models for space-time referring and spatiotemporal reasoning (\textit{ref.}~Table~\ref{tab:res_region_cap}, ~\ref{tab:res_region_qa}, and~\ref{tab:res_more}). 
    }
\label{fig:method}
\end{figure*}

To address the limitations of current video instruction datasets, we introduce \ours (\textit{ref.}~Fig.~\ref{fig:method}), a novel data engine that systematically generates synthetic, fine-grained, spatiotemporally and semantically rich instruction data for training Video LLMs on space-time reference and reasoning tasks. Our approach begins with a modular framework that orchestrates multiple agents—including pre-trained Large Language Models (LLMs), Video LLMs, and Pixel-Level Multimodal Vision Foundation Models (e.g., RexSeek~\cite{rexseek}, GroundingDINO~\cite{liu2023groundingdino} and SAM2~\cite{ren2025grounded_sam2})—to \textit{pseudo-annotate} video metadata with temporally dense and object-centric space-time information. This metadata captures detailed spatial and temporal structures, such as subjects, objects, their locations as masklets (segmentation masks tracked over time), and 
action timelines.
Building on this structured metadata, we leverage in-context learning and well-defined task schemas to guide LLMs in generating high-utility instruction data for tuning Video LLMs.

Unlike existing datasets and synthesis approaches, which often rely on legacy annotations~\cite{li2023videochat,maaz2023video,li2024mvbench,munasinghe2025videoglamm,heo2025omni,sun2025sama} or 
limited-scale but high-cost
human labeling~\cite{athar2025vicas}, \ours automatically produces instruction-response pairs grounded in object-centric, spatiotemporal video structures at scale. Furthermore, our framework supports the generation of multimodal user prompts that mimic realistic human-AI interactions in dynamic environments. These prompts encourage models to reason about spatial references\footnote{Spatial visual prompts may be derived from user interactions (e.g., clicking) or from user gestures, including finger pointing and eye gaze.}~\cite{mani2020point} (e.g., single frame or trajectories of masks\footnote{To support diverse, free-form spatial reference from users (e.g., points, scribbles, and boxes), we choose segmentation masks as the model input because masks contain richer information and many forms of spatial reference can be easily transformed into masks using off-the-shelf tools like SAM2~\cite{sam2}.}) and temporal dynamics (e.g., event sequences, temporal dependencies, and specific time-based moment anchoring). Crucially, our system is designed to handle complex scenarios that challenge existing data synthesis methods—such as scenes containing multiple entities of the same category and 
cases where entities 
do not appear in the first frame, or 
temporarily exit and re-enter the frame—and is capable of processing minutes-long videos rather than just short clips a few seconds long,
thereby fostering more robust and generalizable spatiotemporal perception and reasoning in Video LLMs.

Through the development of \ourseos, we make the following key contributions: 
\begin{itemize}
    \item We provide a scalable methodology for pseudo-labeling videos with temporally dense, object-centric, structured space-time metadata, as well as for generating synthetic instruction-response pairs that promote nuanced spatial and temporal perception and reasoning over videos.
    \item We operationalize data synthesis for 
    an underexplored but essential capability of real-world AI agents—fine-grained, spatiotemporal, object-centric referencing in video-based queries—paving the way for more perceptually grounded and context-aware interactions.
    \item We demonstrate that Video LLMs trained on our data outperform  baselines in tasks requiring spatial and temporal disambiguation. These models also show improved space-time-aware reasoning,
    marking a crucial step toward AI systems that can understand and act within the full space-time fabric of our visual world.
\end{itemize}

\begin{figure*}[t]
	\centering
	\vspace{-20pt}
    \includegraphics[scale=0.80]{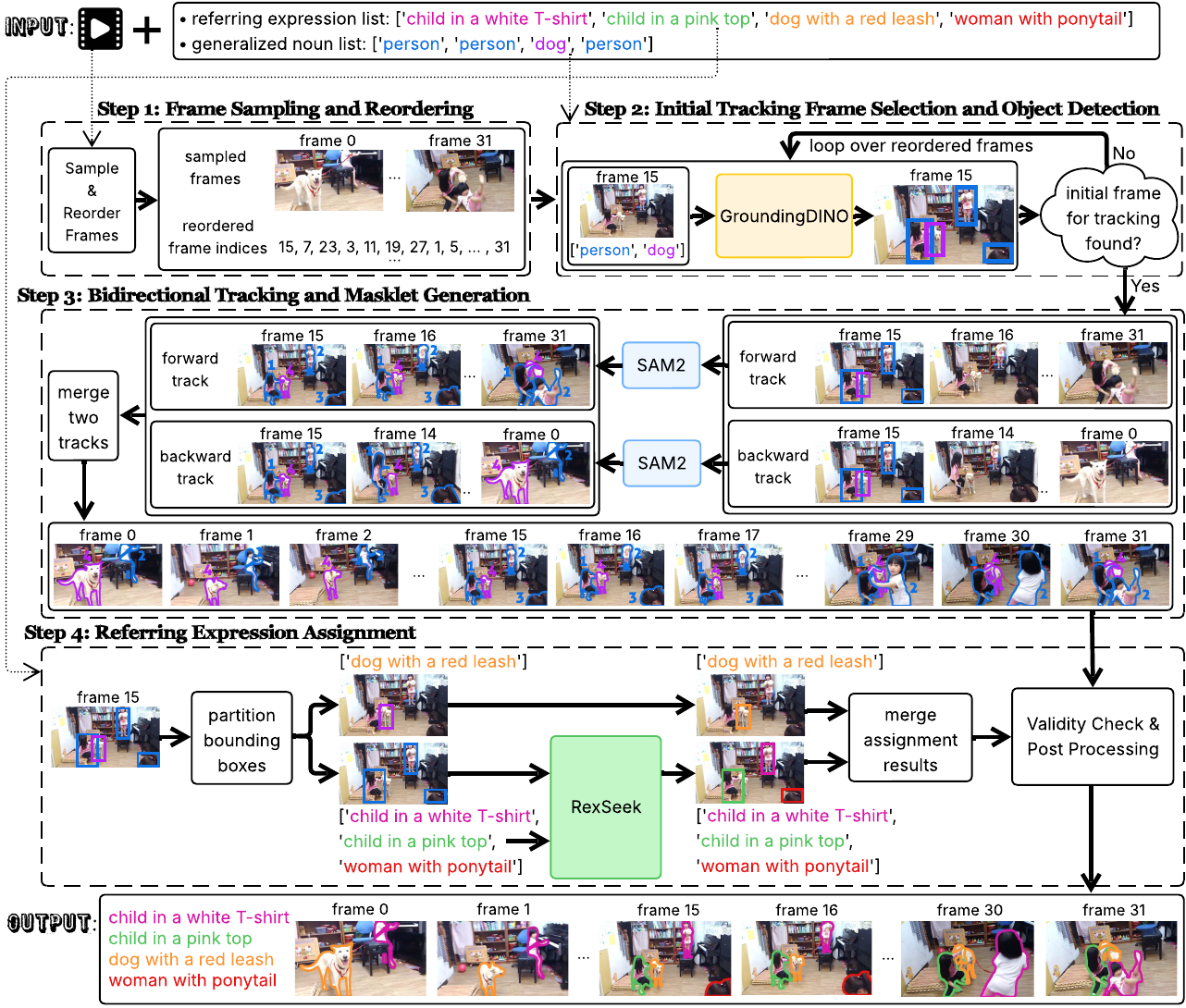}
	\caption{\textbf{Overview of the Referring Masklet Generation Pipeline within \ourseos.} 
    This pipeline produces tracked segmentation masks  from videos with complex structures based on multi-word natural language referring expressions. 
    Our masklet generator 
    is carefully crafted to address key limitations overlooked by prior works~\cite{yuan2025videorefer,munasinghe2023pg,kazakos2025large}
by orchestrating complementary strengths of the state-of-the-art pixel-level vision foundation models to achieve more effective results.
    It robustly handles challenging scenarios, including multiple same- or similar-category entities described differently, entities absent in the first frame, and entities that temporarily exit and re-enter the scene. 
    }
\label{fig:r2m}
\end{figure*}

\section{Methodology}
\label{sec:method}
\vspace{-5pt}

In the following, Sec.~\ref{subsec:auto_data_engine} details the data synthesis process, Sec.~\ref{subsec:data_char} outlines the resulting dataset, and Sec.~\ref{subsec:model_archi} presents modeling choices (w or w/o model architectural changes)
for space-time referring in general-purpose Video LLMs.

\subsection{\ourseos: Automatic Data Engine}
\label{subsec:auto_data_engine}
\vspace{-3pt}

Figure~\ref{fig:method} illustrates the \ours framework for video pseudo-annotation and instruction data generation. All components in the \ours framework are pre-trained, frozen, open-source models, and used off-the-shelf.

\subsubsection{Entity Recognizer}
\vspace{-3pt}
Given an input video, a Video LLM is employed as an \textit{Active Entity Recognizer}, tasked with identifying all active entities present throughout the video. Here, an entity refers to a person, an animal, or an object, while an active entity is defined as any entity that exhibits dynamic behavior—such as movement, interaction with other objects, or performing actions. To guide the model's recognition process, we explicitly provide the definition of active entity in the prompt (see Appendix). We also instruct the model to use language that can clearly distinguish each 
entity from others.

\subsubsection{Referring Parser}
\vspace{-3pt}

Since the Active Entity Recognizer outputs a descriptive paragraph, we use an LLM to extract three structured lists: (1) entity-referring expressions (e.g., ``girl in a white dress''), (2) their noun categories (e.g., ``girl''), and (3) generalized categories (e.g., ``girl'' → ``person'', ``parrot'' → ``bird''). 
Shortening referring expressions and extracting their broader concepts benefits the subsequent referring masklet generation module.

\vspace{-3pt}
\subsubsection{Referring Masklet Generator}
\vspace{-3pt}
The entity-referring expressions and their generalized categories,
together with the video input, are passed to the \textit{Referring Masklet Generator} (Fig.~\ref{fig:r2m}) which generates masklets corresponding to each referring expression.

\vspace{3pt}
\noindent \textit{\textbf{Challenges}:}
Prior methods such as GroundedSAM2~\cite{ren2025grounded_sam2} face several significant challenges when applied to the task of referring masklet generation in complex videos. While GroundedSAM2 supports both image and video input for referring segmentation and tracking, it struggles with multi-word referring expressions—especially in scenes with multiple entities sharing the same noun. This limitation stems from its use of spaCy~\cite{vasiliev2020natural}
to extract a short
noun phrase from complex expressions, which is then used as a prompt for GroundingDINO~\cite{liu2023groundingdino}, an open-vocabulary object detector that expects simple object nouns. Once the bounding boxes of simple object nouns on the first frame are identified, GroundedSAM2 employs SAM2~\cite{ravi2024sam2} to propagate boxes through subsequent frames of the video and generate segmentation masklets.

In contrast, RexSeek~\cite{rexseek} is recently introduced which handles complex referring expressions and performs well in disambiguating entities with similar object nouns. However, it is limited to static images and cannot manage video-specific challenges such as motion blur, occlusion, or object re-identification across frames.

Moreover, GroundedSAM2's performance degrades significantly when the target object is not visible in the first frame. Although SAM2 can handle scenarios where objects temporarily disappear and reappear, it assumes that tracking begins from an ideal, user-provided starting frame in which all target objects are visible. Consequently, selecting a suitable starting frame is critical for achieving effective tracking performance with SAM2. However, GroundedSAM2 automatically initializes SAM2's tracking from the video's first frame, without any mechanism to select a more appropriate frame or to handle complex video conditions.

\noindent \textit{\textbf{Our Referring Masklet Generator}:}
To address these challenges, we introduce a modular pipeline (Fig.~\ref{fig:r2m}).

\noindent \textbf{\texttt{\colorbox{LightPink!20}{Step 1}}: Frame Sampling and Reordering.} The video is sampled into frames, which are then reordered using a heuristic that assumes important content typically occurs near the middle of the video. This reordering facilitates a more efficient search for a suitable tracking start frame.

\noindent \textbf{\texttt{\colorbox{LightPink!20}{Step 2}}: Initial Tracking Frame Selection and Object Detection.} This step identifies an initial frame for tracking and simultaneously performs object detection using generalized entity nouns. Our key finding is that GroundingDINO performs more reliably when prompted with simple nouns (e.g., ``person'' instead of ``bride''), and therefore, we use generalized entity nouns as prompts. The selected starting frame is the first in the reordered frame list where the number of detected objects matches or exceeds the number of referring expressions. If no such frame exists, the frame with the highest number of detected objects is used.

\noindent \textbf{\texttt{\colorbox{LightPink!20}{Step 3}}: Bidirectional Tracking and Masklet Generation.} Using SAM2, tracking is performed both forward and backward from the selected initial frame, with object bounding boxes of generalized nouns detected by GroundingDINO on that frame as the input prompt. The resulting mask tracks from the two directions are then merged based on overlapping detections in the initial frame. 
Each generalized noun may have one or multiple masklets produced, depending on the specific video scenario.

\noindent \textbf{\texttt{\colorbox{LightPink!20}{Step 4}}: Referring Expression Assignment.} RexSeek is used to assign each referring expression to its corresponding masklet.
For nouns with multiple candidate masklets (e.g., ``person'' referring to several individuals in the video), the full referring expressions are leveraged to disambiguate and establish the correct associations. This assignment is performed on the same frame used to initialize segmentation and tracking. Specifically, we first group masklets by their associated generalized nouns. Then, for each group requiring association resolution, given the frame and the bounding boxes of that noun in the frame, we prompt RexSeek with the prompt template: ``\texttt{Please detect \{referring\} in this image. Answer the question with object indexes.}'' 
The full referring expression replaces the word \texttt{\{referring\}}  in the prompt template.

\subsubsection{Video Clipper}

To capture temporally dense dynamics of each entity (e.g., movements, actions, and their coarse temporal boundaries), we segment the video into short clips using visual scene changes and semantic frame-level shifts. We first apply PySceneDetect’s ContentDetector~\cite{scenedetect2025} with a threshold of $20$, empirically chosen based on qualitative assessment.

Many videos remained unclipped by PySceneDetect, despite clear action or event transitions, due to its reliance on HSV-based frame differences that miss semantic changes. To overcome this, we extract SigLIP~\cite{zhai2023sigmoid} frame embeddings at $3$ FPS and apply a custom clipping algorithm using Hierarchical Agglomerative Clustering to segment semantically distinct video segments.

\begin{figure}[t]
	\centering
    \includegraphics[scale=0.58]{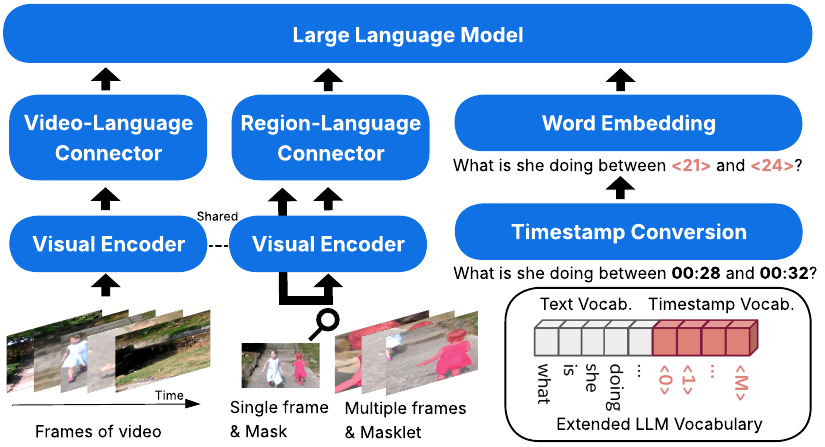}
	\caption{\textbf{Model Architecture}: Plug-and-play modules (Region-Language Connector, Timestamp Conversion) enhance general-purpose Video LLMs with space-time referring capabilities. It is worth noting
that incorporating these modules for space-time referring
is not strictly necessary (see Sec.~\ref{subsec:model_archi}).
    }
\label{fig:model_architecture}
\vspace{-5pt}
\end{figure}

\subsubsection{Video Transcriber}

The \textit{Video Transcriber} is a Video LLM that generates behavior-centric descriptions for entities, clip by clip. For each entity in the referring expression list, we iterate through the video clips and apply a two-step prompting process:
(1) \textit{Presence Check}: We first ask the model, ``Is there a \texttt{entity}? Answer `Yes' or `No'.''
(2) \textit{Behavior Description}: If the model responds with ``Yes'', we follow up with: ``What clearly happens to the \texttt{entity}? Describe only what is visibly happening to the \texttt{entity}, without inference or assumptions.''
The term ``\texttt{entity}'' is substituted with the actual referring expression (e.g., ``girl in a white dress''). 

This method yields a structured, temporally-dense video transcript focused on explicit, observable dynamics of individual entities.

\subsubsection{Video Instruction Data Generator}
Finally, using the video metadata we have collected—such as structured video transcripts and language-described masklets—we generate instruction-style question-answer (QA) pairs either through templates or LLMs.

For example, given temporally-dense descriptions of entities across clips and knowledge of when each entity first appears, we can design templates to generate questions that test the understanding of temporal ordering—e.g., identifying the sequence in which entities appear. We can also ask about the actions of a specific entity within a given time interval, even if the entity does not appear during that interval but is present in other segments of the video. 

Template-based question-answer generation typically involves designing a separate template for each task type (e.g., entity behavior captioning, temporal ordering). Crafting templates that effectively capture long-range dependencies—and implementing code to automatically generate QA pairs that conform to these templates—is difficult to scale. Alternatively, we can use the video transcript along with a set of in-context examples, and leverage an LLM to generate QA pairs that follow our predefined task definitions.

Moreover, to encourage reliance on masklets for resolving references, an LLM uses the entity referring expression list (which includes entities grounded with masklets) to replace specific language-based entity references with pronouns or generic terms (e.g., ``she'', ``he'', ``the person'') in the generated questions.

We present the question task types and their definitions, used in both the template-based and LLM-based approaches, in Table~\ref{tab:question_types} of the Appendix. Each task is carefully designed to focus on space-time perception and reasoning.

\subsection{Synthesized Data}
\label{subsec:data_char}

\begin{figure}[t]
	\centering
    \includegraphics[scale=0.32]{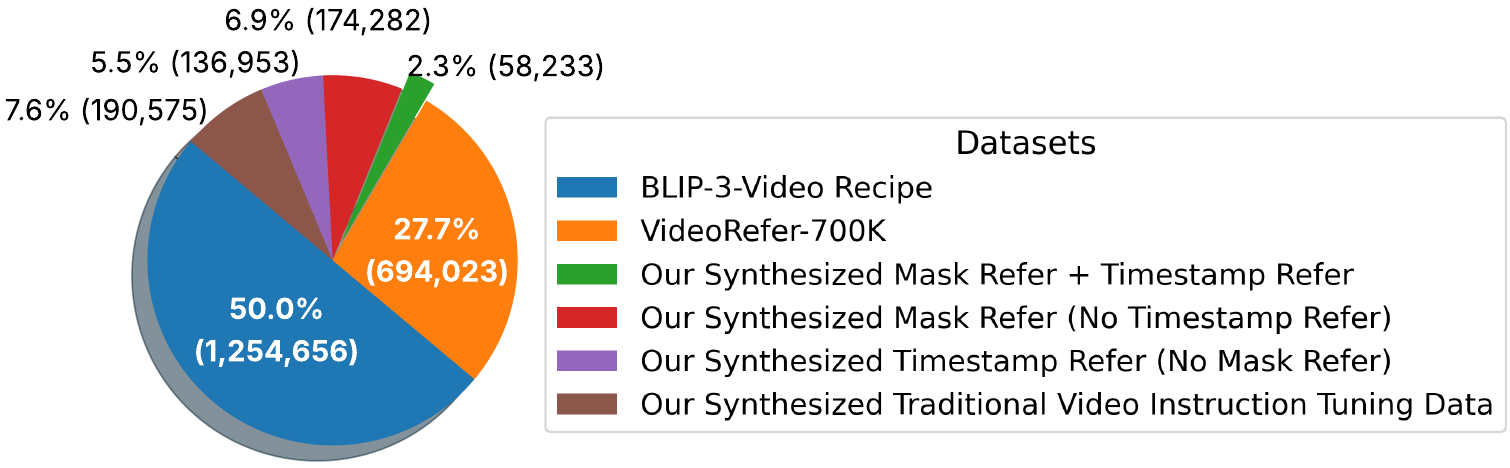}
    \vspace{-1pt}
	\caption{Data composition of our \colorbox{LightSkyBlue!20}{final recipe} used in our experiments in Sec.~\ref{sec:exp}.
    }
\label{fig:data_composition}
\end{figure}

We implement \ours leveraging Tarsier-34b~\cite{wang2024tarsier}
and Qwen2.5-32B-Instruct~\cite{qwen2.5} as the Video LLM and LLM resp. For the Masklet Generator, we use RexSeek-3B~\cite{rexseek}, SAM2 (sam2.1\_hiera\_large)~\cite{ravi2024sam2} 
and GroundingDINO (grounding-dino-tiny)~\cite{liu2023groundingdino}. By the time of writing, we have generated video metadata and 
$947,854$
instruction-response pairs using $4,253$ NExT-QA~\cite{xiao2021next} videos (excluding test set; avg. $40$ seconds long and up to $3$ minutes long). 
We illustrate the data composition of our final recipe in Fig.~\ref{fig:data_composition}.
Our synthesized instruction data is organized into $\mathbf{8}$ distinct groups based on whether masks or timestamps appear in the question and the QA-generation source (template or LLM), covering $\mathbf{11}$ different question task types. For details, please refer to Table~\ref{tab:data-groups} and Table~\ref{tab:question_types} in the Appendix.

\subsection{Modeling for Space-Time Referring}
\label{subsec:model_archi}
\noindent \textbf{Modeling with Architectural Enhancements}. We add plug-in-and-play modules to established general-purpose Video LLMs to unleash their fine-grained mask-level comprehension at any specific regions and any timestamps for a given video. To support detailed region-level understanding, we incorporate the spatiotemporal object encoder design from VideoRefer~\cite{yuan2025videorefer}, which enables the model to understand fine-grained mask and masklet. For precise timestamp-level comprehension, we introduce learning special temporal tokens inspired by GroundedLLM~\cite{wang2024grounded}, allowing the model to interpret specific moments in time properly. For details, please refer to Appendix.
The resulting model (Fig.~\ref{fig:model_architecture}) is a next-token-prediction Video LLM with fine-grained, mask-level comprehension across arbitrary spatial regions and temporal segments of a video.

\vspace{5pt}
\noindent \textbf{Modeling without Architectural Changes}. It is worth noting that incorporating these modules for space-time referring is not strictly necessary. 
We also explore visual prompting approaches—SoM~\cite{yang2023set} for masklet comprehension and NumberIt~\cite{wu2025number} for timestamp understanding.

\section{Evaluation Details}
\label{sec:exp_setup}
\subsection{Training Details}

To evaluate the quality of our synthesized instruction data, we integrate it into a base video instruction tuning recipe, which combines the video instruction-tuning data used by BLIP-3-Video~\cite{blip3video} with VideoRefer-700K~\cite{yuan2025videorefer}. Our baseline is the model tuned on this base recipe.
The video instruction-tuning data used by BLIP-3-Video comprises data
from multiple sources, including Mira~\cite{ju2024miradata}, VideoInstruct-100K~\cite{maaz2023video}, MSVD-QA~\cite{xu2017video}, MSRVTT-QA~\cite{xu2017video}, ActivityNet-QA~\cite{yu2019activitynet}, TGIF-QA~\cite{jang2017tgif}, and NExT-QA~\cite{xiao2021next}.
VideoRefer-700K is a recently released instruction-tuning dataset for video mask and masklet referring tasks, but it lacks timestamp-referring instructions.

The full model 
is tuned except the visual encoder. The visual encoder is not tuned due to insufficient data, which prevents effective tuning. Since the encoder is designed to extract complex visual patterns and features from raw RGB signals, it requires a large amount of data to generalize well. 
To be specific, we start from an image-comprehension vision LLM, the pre-trained BLIP-3~\cite{blip3} model, with additional untrained architectural enhancements from BLIP-3-Video~\cite{blip3video}, as well as those described in Sec.~\ref{subsec:model_archi}. We adapt BLIP-3 for video and masklet comprehension by fine-tuning the full model illustrated in Fig.~\ref{fig:model_architecture}, except for the visual encoder, using $32$ frames per video and $32$ temporal tokens.
Other hyperparameters, such as learning rate and batch size, were selected based on downstream evaluation results for the `Baseline' model presented in the result tables. However, when tuning the model using recipes integrated with our data, we did not change any hyperparameters from those used in the `Baseline' model.
The training takes roughly $1$ day and requires $3\times 8$ H200 GPUs. 
The resulting model has
$4$B parameters.

In the baseline and ablation models, if its training data lacks mask-referring instructions, the corresponding modules are excluded; likewise, timestamp-related modules are omitted if timestamp-referring instructions are not present in training.
Therefore, the `Baseline Ablation' model presented in the result tables shares the same architecture as BLIP-3-Video~\cite{blip3video}; the model does not include the plug-and-play modules described in Sec.~\ref{subsec:model_archi}, as its training data lacks instructions that refer to masks or timestamps. The `Baseline' model does not have `Timestamp Conversion' or an extended LLM vocabulary for learning special temporal tokens (see Fig.~\ref{fig:model_architecture}) due to the lack of instruction data involving specific timestamps.

Our synthetic data includes full-length masklets per referring entity, but for efficient training, we sample a single mask on a random frame per instruction-response pair. At evaluation, we use the full masklet. Training with full masklets is expected to further improve performance.

\subsection{Benchmark Details}
\label{subsec:eval_details}

We describe the evaluation benchmarks for Mask-Referred Regional Description, Mask-Referred Regional QA, and Timestamp-Referred Video QA below, as these represent less common evaluation settings for Video LLMs.

\noindent \textbf{VideoRefer-Bench\textsuperscript{D}~\cite{yuan2025videorefer}}
assesses the model's ability to describe an
entity across a video, given a mask or masklet of that entity. The benchmark comprises $400$ videos from the test set of Panda-70M~\cite{chen2024panda}. 

To evaluate performance on this benchmark, we 
use the following instruction template: ``\texttt{Please give a detailed description of the highlighted object <region> in the video.}'' The word \texttt{<region>} is substituted with model-extracted regional tokens if the model has built-in mechanisms to extract regional features.

The model evaluation is performed by GPT-4o by assigning scores to the generated predictions on a scale range from $0$ to $5$ across the following four dimensions~\cite{yuan2025videorefer}:
\begin{itemize}
    \item Subject Correspondence: This dimension evaluates whether the subject of the generated description accurately corresponds to that specified in the ground truth.

    \item Temporal Description: This aspect analyzes whether the representation of the object’s motion is consistent with the actual movements.
    
    \item Appearance Description: This criterion assesses the accuracy of appearance-related details, including color, shape, texture, and other relevant visual attributes.
    
    \item Hallucination Detection: This facet identifies discrepancies by determining if the generated description includes any facts, actions, or elements absent from reality, like imaginative interpretations or incorrect inferences.
\end{itemize}

\noindent \textbf{VideoRefer-Bench\textsuperscript{Q}~\cite{yuan2025videorefer}} evaluates a model’s ability to answer video entity-related questions, given one or more entities' masks or masklets within a video. The benchmark includes $1,000$ multiple-choice questions spanning $198$ videos sourced from various datasets, including 
the test set of MeViS~\cite{MeViS}, A2D-Sentences~\cite{gavrilyuk2018actor}, and Refer‑YouTube‑VOS~\cite{wu2022language}. Questions are crafted to assess different dimensions of understanding, including Basic Questions, Sequential Questions, Relationship Questions, Future Predictions and Complex/Reasoning Questions.

Sequential Questions typically ask about entity action and ordering; Basic Questions typically concern attributes like object color. Relationship Questions involve more than one object regions in the question. Future Predictions involve weakly grounded reasoning about forthcoming events.
Notably, models generally perform best on Complex/Reasoning Questions, making this category the easiest despite its name.

We use the following instruction template: ``\texttt{Please answer the following question about the <region>. \{question\}}''.

\noindent \textbf{Timestamp-based Yes/No QA on QVHighlights} is a task that repurposes existing annotations from the video highlight detection dataset, QVHighlights~\cite{qvhighlights}. Specifically, for each annotated segment—defined by a start and end timestamp and an associated language description—we construct a question prompt in the following form: \\
`\texttt{Does the following description accurately reflect what happens in the video between <start\_time> and <end\_time>? Description: \{description\}. Respond with `Yes' or `No' only.}'' 
 Each of these prompts is assigned the ground truth answer ``\texttt{Yes}''. 

To generate negative (i.e., ``\texttt{No}'') samples, we randomly select segments from the same video that do not overlap with any annotated intervals. To ensure this, we first expand each annotated timestamp by a buffer of $5$ seconds on both sides, then merge overlapping intervals to form a set of excluded ranges. We then identify all remaining gaps in the video timeline that lie outside these excluded regions. From these valid gaps, we randomly select a new segment that satisfies a minimum duration of $10$ seconds. A description from an annotated segment is then paired with this unrelated time window to form a mismatched QA example with the correct answer ``\texttt{No}''.

We ensure a balanced answer distribution, with almost $50\%$ of the samples labeled as ``\texttt{Yes}'' and $50\%$ as ``\texttt{No}''.

For each question, we substituted \texttt{<start\_time>} and \texttt{<end\_time>} with their corresponding timestamps.
For models that do not learn temporal tokens, timestamps are represented by default in the \texttt{HH:MM:SS.xxx} format.
For models that do learn temporal tokens, we use temporal tokens to substitute \texttt{<start\_time>} and \texttt{<end\_time>}. 
For example, if a model learns $32$ temporal tokens and the video's duration is 90 seconds, a timestamp like \texttt{00:00:19.228} is converted to \texttt{$<$7$>$} (\(\tfrac{19.228}{90} \times 32 \approx 7\)).

\section{Experiments}
\label{sec:exp}

\subsection{Quantitative Results}

\subsubsection{Main Results and Findings}
We explore the final data recipe and ablate groups of our synthesized instruction data, and results are listed in Tables~\ref{tab:res_region_cap},~\ref{tab:res_region_qa}, and~\ref{tab:res_more}.
We created model variants by incrementally adding our data groups to the base recipe. These groups were categorized based on whether the instructions involved mask- or timestamp-referring cues, and whether the QA pairs were generated by an LLM or derived from templates.
We summarize our findings as follows:

\vspace{3pt}
\noindent \textbf{\textit{Finding 1}: The video space-time referring task, which requires understanding both the full video and the user-specified regions at specific times, enhances spatiotemporal understanding in Video LLMs.}

Integrating the base recipe with our LLM-synthesized $\mathcal{G}7$ data—featuring queries with mask and timestamp references—boosts performance across several tasks: mask-referred video regional description improves from $3.28$ to $3.34$, mask-referred video QA from $0.665$ to $0.672$, timestamp-based QA from $0.5288$ to $0.5390$, and TempCompass~\cite{liu2024tempcompass} from $60.100$ to $60.650$. Impressively, $\mathcal{G}7$ contains just $27$K samples ($1.39$\% of the base recipe).

Incorporating additional data containing queries with only mask references ($\mathcal{G}6$) leads to consistent performance gains across all tasks—except for QVHighlights~\cite{qvhighlights}. On this benchmark, using both $\mathcal{G}6$ and $\mathcal{G}7$ results in a slightly lower score ($0.5337$) compared to using only $\mathcal{G}7$ ($0.5390$). We hypothesize that this drop is due to the relatively smaller proportion of timestamp-referred queries, as the inclusion of $174$K mask-referred instances may dilute the supervision of timestamp-based video understanding.
Despite this, the combined data still outperforms the baseline, which lacked any timestamp-referred training examples. Notably, further augmenting training with $\mathcal{G}6$ yields the highest score ($4.4525$) on the Subject Correspondence task, which evaluates
the model's ability to align the described subject with that specified in the ground truth,
highlighting the benefits of our mask-referring data in enhancing model understanding of specific regions within the video content.

To further enhance the ability of model to interpret timestamp-based video content, we introduce $\mathcal{G}8$, which yields a QVHighlights score of $0.5672$. $\mathcal{G}8$ combines timestamp and mask references and is derived from a subset of $\mathcal{G}5$, a template-generated timestamp referring data. We sampled remaining $\mathcal{G}5$ that were not converted into mask referring data and incorporate it into the training recipe, performance on QVHighlights improved into $0.5900$. 

\vspace{3pt}
\noindent \textbf{\textit{Finding 2}: Traditional video instruction data—when designed with a focus on dynamics and enriched with both positive and negative questions—can further improve performance, even on space-time referring tasks.}

Our final recipe incorporates an additional component, $\mathcal{G}1$, which is generated from templates and specifically designed to focus on entity behavior. It includes negative questions—such as asking the model to describe the behavior of entities that are absent from a given video segment—to help the model avoid relying on shortcuts or making incorrect assumptions.
Surprisingly, this addition not only enhances performance on video temporal perception and reasoning tasks (TempCompass and VideoMME~\cite{videomme}), but also improves results on space-time referring understanding tasks:
Compared to the baseline, the average score on mask-referred video regional description increases to $3.39$; mask-referred video QA accuracy improves to $0.688$; timestamp-based QA rises to $0.6031$; performance on TempCompass reaches $61.675$; and performance on VideoMME improves to $37.70$.  
The final recipe consistently and significantly improves performance across benchmarks.

\vspace{3pt}
\noindent \textbf{\textit{Finding 3}: Template-generated event sequencing data may enhance higher-level, long-term temporal reasoning skills, but this can come at the cost of precise, fine-grained spatial-temporal understanding. Therefore, a more balanced data mixture may be necessary to support both broad temporal abstractions and localized, detailed spatial-temporal comprehension.}

$\mathcal{G}2$ comprises multiple-choice traditional video instruction data that tasks the model with identifying the correct temporal order from the wrong ones in which entities first appear in the video. We hypothesize that this task format fosters long-range temporal reasoning abilities. Empirically, \textit{before incorporating $\mathcal{G}1$}, we find that incorporating $\mathcal{G}2$ improves performance on temporally-focused benchmarks such as TempCompass with short videos (from $60.750$ to $60.925$), as well as VideoMME-a long-video understanding benchmark containing hour-long videos (from $35.90$ to $41.65$), supporting this hypothesis.

However, we observe a performance drop when $\mathcal{G}2$ is added in tasks like mask-referred captioning, mask-referred QA, and timestamp-referred QA. We attribute this decline to a trade-off in the type of reasoning the model develops: while $\mathcal{G}2$ promotes higher-level long-term temporal reasoning skills, the mask or timestamp-referred understanding tasks demand fine-grained, localized understanding of video content. Thus, the broader temporal abstractions encouraged by $\mathcal{G}2$ may come at the expense of precision in spatial-temporal fine-grained, localized understanding.

After incorporating $\mathcal{G}1$, further adding $\mathcal{G}2$ leads to performance drop across all benchmarks, even though $\mathcal{G}2$ contains only $1$K samples. This suggests that $\mathcal{G}1$, which incorporates negative questions and dynamic understanding, may already cultivate a certain degree of temporal reasoning. Consequently, the benefits of $\mathcal{G}2$ may be diminished or overshadowed in the presence of $\mathcal{G}1$.

\begin{table*}[t]
\setlength{\tabcolsep}{2.8pt}
  \aboverulesep=0ex
  \belowrulesep=0ex 
\centering
\footnotesize
\vspace{-5pt}
\begin{tabular}{c p{8.8cm} c c r}
\toprule
\textbf{Group} & \textbf{Description} & \textbf{Visual Input} & \textbf{Task Types (\textit{ref.} Tab.~\ref{tab:question_types})} & \textbf{\# Samples} \\
\midrule
$\mathcal{G}1$ & Traditional data, template generated  & Clip & $1$,$2$,$3$,$4$ & $190,575$ \\
$\mathcal{G}2$ & Traditional data, template generated  & Video & $5$ & $1,316$ \\
$\mathcal{G}3$ & Traditional data, LLM generated & Video & $7$ & $288,630$ \\
$\mathcal{G}4$ & \textbf{Timestamp referring} data, LLM generated & Video & $8$,$9$,$10$,$11$ & $44,243$ \\
$\mathcal{G}5$ & \textbf{Timestamp referring} data, template generated & Video & $6$ & $190,575$ \\
$\mathcal{G}6$ & \textbf{Mask referring} instead of language referring, derived from $\mathcal{G}3$ & Video & $7$ & $174,282$ \\
$\mathcal{G}7$ & \textbf{Mask referring} instead of language referring, derived from $\mathcal{G}4$ & Video & $8$,$9$,$10$,$11$ & $27,092$ \\
$\mathcal{G}8$ & \textbf{Mask referring} instead of language referring, derived $\mathcal{G}5$ & Video & $6$ & $31,141$ \\
\bottomrule
\end{tabular}
\vspace{-8pt}
\caption{\textbf{Overview of our synthesized instruction-tuning data}.
Using \ourseos, we generated $947,854$
instruction data from only $4,253$ NExT-QA videos.
Our synthesized data includes a diverse range of temporally focused questions (i.e., tasks)—such as temporal relations, event sequencing, time-referencing, and coarse time localization. It incorporates both positive and negative questions to help the model learn to avoid relying on shortcuts or making incorrect assumptions.
Although our \colorbox{LightSkyBlue!20}{final recipe} introduced only $545$ additional videos beyond the baseline, it yielded noticeably improved performance across multiple benchmarks (\textit{cf.} Tab.~\ref{tab:res_region_cap}, Tab.~\ref{tab:res_region_qa}, and Tab.~\ref{tab:res_more}).}
\vspace{-5pt}
\label{tab:data-groups}
\end{table*}

\begin{table*}[t!]
\setlength{\tabcolsep}{2.5pt}
\aboverulesep=0ex
\belowrulesep=0ex 
\centering
\footnotesize
\begin{tabular}{lrccccc}
\toprule
\multirow{2}{*}{\textbf{Mask-Referred Regional Description} (VideoRefer-Bench\textsuperscript{D}~\cite{yuan2025videorefer})} & \makecell{Samples \\ Added (\%)} & \multirow{2}{*}{\textbf{Avg.}} & \makecell{Subject \\ Correspondence} & \makecell{Temporal \\ Description} & \makecell{Appearance \\ Description} & \makecell{Hallucination \\ Detection} \\
\midrule
GPT-4o   & N/A & $3.25$ & $4.15$ & $3.11$ & $3.31$ & $2.43$ \\
GPT-4o-mini   & N/A & $3.05$ & $3.89$ & $2.62$ & $3.18$ & $2.50$ \\
Baseline Ablation: Video Instruction-Tuning Data~\cite{blip3video}  & N/A   & $2.7308$ & $3.5200$ & $2.4235$ & $2.5639$ & $2.4160$ \\ 
\rowcolor{LightGray!15}  Baseline: Base Recipe (1,948,679 samples)  & N/A & $3.2837$ & $4.3775$ & $2.9523$ & $3.1075$ & $2.6975$ \\ \midrule  
\rowcolor{LightSkyBlue!20} \quad   + $\mathcal{G}6$ + $\mathcal{G}7$  + $\mathcal{G}8$  + Remaining $\mathcal{G}5$ (Sampled) + $\mathcal{G}1$ & $28.73\%$  & $\mathbf{3.3947}$ & $\underline{{4.4400}}$ & $3.0575$ & $\mathbf{3.2763}$ & $2.8050$ \\
\quad   + $\mathcal{G}6$ + $\mathcal{G}7$  + $\mathcal{G}8$  + Remaining $\mathcal{G}5$ (Sampled) + $\mathcal{G}2$  & $19.02\%$ & $3.3537$ & $4.3650$ & $3.0150$ & $\underline{{3.2675}}$ & $2.7675$ \\
\quad   + $\mathcal{G}6$ + $\mathcal{G}7$  + $\mathcal{G}8$  + Remaining $\mathcal{G}5$ (Sampled) + $\mathcal{G}1$ + $\mathcal{G}2$ & $28.80\%$ &  $3.3742$ & $4.3975$ & $\underline{{3.0676}}$ & $3.2493$ & $2.7825$ \\
\quad   + $\mathcal{G}6$ + $\mathcal{G}7$  + $\mathcal{G}8$  + Remaining $\mathcal{G}5$ (Sampled)  & $18.95\%$ & $3.3740$ & $4.4025$ & $3.0300$ & $3.2512$ & $\underline{{2.8125}}$ \\
\quad   + $\mathcal{G}6$ + $\mathcal{G}7$  + $\mathcal{G}8$  & $11.93\%$ & $\underline{{3.3821}}$ & $4.3625$ & $\mathbf{3.1150}$ & $3.2317$ & $\mathbf{2.8195}$ \\
\quad   + $\mathcal{G}6$ + $\mathcal{G}7$  & $10.33\%$ & $3.3710$ & $\mathbf{4.4525}$ & $3.0579$ & $3.2412$ & $2.7325$ \\
\quad   + $\mathcal{G}7$  & $1.39\%$ & $3.3421$ & $4.3825$ & $3.0150$ & $3.2311$ & $2.7400$ \\
\bottomrule
\end{tabular}
\vspace{-8pt}
\caption{\textbf{Regional Description} results on VideoRefer-Bench\textsuperscript{D} (\textbf{Best}/\underline{{Second}}). This evaluation assesses the model's ability to describe an entity across a video, given a mask or masklet of that entity. If a video contains multiple entities, the model must leverage the region input to distinguish and describe the correct one. 
Across result tables: The \colorbox{LightSkyBlue!20}{blue-highlighted} row represents our final recipe that performs well across all benchmarks. GPT-4o and GPT-4o-mini results are listed but excluded when highlighting the top two performers.
}
\vspace{-5pt}
\label{tab:res_region_cap}
\end{table*}

\begin{table*}[t!]
\setlength{\tabcolsep}{3.8pt}
\aboverulesep=0ex
\belowrulesep=0ex 
\centering
\footnotesize
\begin{tabular}{lrcccccc}
\toprule
\multirow{2}{*}{\textbf{Mask-Referred  Regional QA} (VideoRefer-Bench\textsuperscript{Q}~\cite{yuan2025videorefer})}  & \makecell{Samples \\ Added (\%)} & \multirow{2}{*}{\textbf{Avg.}} & \makecell{Sequential \\ Questions} & \makecell{Relationship \\ Questions} & \makecell{Basic \\ Questions} & \makecell{Future \\ Predictions} & \makecell{Reasoning \\ Questions} \\
\midrule
GPT-4o  & N/A & $0.713$  & $0.745$  &  $0.660$ &  $0.623$ & $0.737$  & $0.880$  \\ 
GPT-4o-mini  & N/A & $0.658$  &  $0.671$ & $0.565$ &  $0.576$ &  $0.754$ &  $0.859$ \\  
Baseline Ablation: Video Instruction-Tuning Data~\cite{blip3video}  & N/A & $0.621$ & $0.6015$ & $0.4920$ & $0.6042$ & $0.7192$ & $0.8321$ \\ 
\rowcolor{LightGray!15}  Baseline: Base Recipe (1,948,679 samples)  & N/A & $0.665$ & $0.6367$ & $0.5476$ & $0.6382$ & $0.7807$ & $0.8741$ \\ \midrule
\rowcolor{LightSkyBlue!20} \quad   + $\mathcal{G}6$ + $\mathcal{G}7$  + $\mathcal{G}8$  + Remaining $\mathcal{G}5$ (Sampled) + $\mathcal{G}1$  &  $28.73\%$ & $\mathbf{0.688}$ & $\mathbf{0.6640}$ & $\underline{{0.5753}}$  & $0.6680$ & $\mathbf{0.8070}$ & $0.8671$ \\ 
\quad   + $\mathcal{G}6$ + $\mathcal{G}7$  + $\mathcal{G}8$  + Remaining $\mathcal{G}5$ (Sampled)  + $\mathcal{G}2$  & $19.02\%$ & $0.677$ & $0.6250$ & $0.5634$ & $0.6808$ & $0.7894$ & $\underline{{0.8741}}$ \\ 
 \quad   + $\mathcal{G}6$ + $\mathcal{G}7$  + $\mathcal{G}8$  + Remaining $\mathcal{G}5$ (Sampled) + $\mathcal{G}1$ + $\mathcal{G}2$  & $28.80\%$  & $0.683$ & $0.6523$ & $0.5555$ & $\mathbf{0.6936}$ & $\underline{{0.7982}}$ & $0.8531$ \\ 
 \quad   + $\mathcal{G}6$ + $\mathcal{G}7$  + $\mathcal{G}8$  + Remaining $\mathcal{G}5$ (Sampled)  &  $18.95\%$ & $0.681$ & $\mathbf{0.6640}$   & $0.5674$ & $0.6510$ & $0.7894$ & $\underline{{0.8741}}$ \\ 
  \quad   + $\mathcal{G}6$ + $\mathcal{G}7$  + $\mathcal{G}8$  & $11.93\%$ & $0.678$ & $0.6289$ & $0.5714$ & $\underline{{0.6851}}$ & $0.7807$ & $0.8601$ \\ 
  \quad   + $\mathcal{G}6$ + $\mathcal{G}7$   & $10.33\%$ & $\underline{{0.685}}$ & $\underline{{0.6601}}$ & $\mathbf{0.5952}$ & $0.6553$ & $0.7543$ & $\mathbf{0.8811}$ \\ 
\quad   + $\mathcal{G}7$  &  $1.39\%$  & $0.672$ & $0.6523$ & $0.5674$ & $0.6425$ & $0.7631$ & $0.8671$ \\ 
\bottomrule
\end{tabular}
\vspace{-8pt}
\caption{\textbf{Regional QA} results on VideoRefer-Bench\textsuperscript{Q}(\textbf{Best}/\underline{{Second}}). This evaluation assesses the model's ability to answer questions about one or more entities in a video, given a mask or masklet of the entity. 
}
\vspace{-5pt}
\label{tab:res_region_qa}
\end{table*}

\begin{table*}[t!]
\setlength{\tabcolsep}{0.3pt}
  \aboverulesep=0ex
  \belowrulesep=0ex 
\centering
\scriptsize
\begin{tabular}{lr|c||ccccc|c}
\toprule
\multirow{2}{*}{\textbf{Timestamp-Referred QA (Yes/No) \& Traditional QA}}  & \multirow{2}{*}{\makecell{Samples \\ Added (\%)}}  & \multirow{2}{*}{\textbf{QVHighlights~\cite{qvhighlights}}} & \multicolumn{5}{c|}{\textbf{TempCompass~\cite{liu2024tempcompass}}} & \multirow{2}{*}{\textbf{VideoMME~\cite{videomme}}} \\
 &  & &  \multirow{1}{*}{\textbf{Avg.}} & \makecell{Yes / No} & \makecell{MCQ} & \makecell{Caption Matching} & \makecell{Captioning} &    \\
\midrule
Baseline Ablation: Video Instruction-Tuning Data~\cite{blip3video}   & N/A  & $0.5297$ & $58.525$ & $57.3$  & $59.3$ & $67.5$ & $50.0$  & $35.85$ \\
\rowcolor{LightGray!15}  Baseline: Base Recipe (1,948,679 samples)  & N/A    & $0.5288$ & $60.100$ & $61.9$ & $58.4$ & $70.5$  & $49.6$ & $37.45$ \\ \midrule
\rowcolor{LightSkyBlue!20} \quad   + $\mathcal{G}6$ + $\mathcal{G}7$  + $\mathcal{G}8$  + Remaining $\mathcal{G}5$ (Sampled) + $\mathcal{G}1$ & $28.73\%$  & $\mathbf{0.6031}$ & $\mathbf{61.675}$ & $\underline{{65.1}}$ & $59.1$ & $70.9$ & $\mathbf{51.6}$ & $37.70$  \\
\quad   + $\mathcal{G}6$ + $\mathcal{G}7$  + $\mathcal{G}8$  + Remaining $\mathcal{G}5$ (Sampled) + $\mathcal{G}2$ & $19.02\%$ & $0.5774$ & $60.925$ & $63.8$ & $58.9$  & $\underline{{71.1}}$ & $49.9$ & $\mathbf{41.65}$ \\
\quad   + $\mathcal{G}6$ + $\mathcal{G}7$  + $\mathcal{G}8$  + Remaining $\mathcal{G}5$ (Sampled) + $\mathcal{G}1$ + $\mathcal{G}2$ & $28.80\%$ & $\underline{{0.5951}}$ & $61.250$ & $64.5$ & $\mathbf{59.8}$ & $70.5$ & $50.2$ & $34.70$  \\
\quad   + $\mathcal{G}6$ + $\mathcal{G}7$  + $\mathcal{G}8$  + Remaining $\mathcal{G}5$ (Sampled) & $18.95\%$  &  $0.5900$ & $60.750$ & $63.4$ & $58.8$ & $70.3$ & $50.5$ & $35.90$  \\
\quad   + $\mathcal{G}6$ + $\mathcal{G}7$  + $\mathcal{G}8$ & $11.93\%$ & $0.5672$ & $\mathbf{61.675}$ & $\mathbf{66.0}$ & $59.3$ & $\mathbf{71.4}$ & $50.0$ & $38.15$  \\
\quad   + $\mathcal{G}6$ + $\mathcal{G}7$ & $10.33\%$  & $0.5337$ & $\underline{{61.550}}$ & $\underline{{65.1}}$ & $\underline{{59.4}}$ & $70.7$ & $\underline{{51.0}}$ & $\underline{{39.15}}$  \\
\quad   + $\mathcal{G}7$ & $1.39\%$ & $0.5390$ & $60.650$ & $63.0$ & $58.7$ & $70.1$ & $50.8$ & $33.90$  \\
\bottomrule
\end{tabular}
\vspace{-8pt}
\caption{\textbf{Results of timestamp-based QA on QVHighlights and traditional QA focusing on  temporal cues using TempCompass and VideoMME} (\textbf{Best}/\underline{{Second}}). 
We transform annotations from QVHighlights into questions tied to specific timestamps, each expecting a `Yes' or `No' answer. For VideoMME, we report results on the Temporal Perception and Temporal Reasoning subsets (no subtitles).
While VideoMME benchmarks long video understanding (up to an hour), our data of entirely short videos can help improve model performance.
} 
\vspace{-5pt}
\label{tab:res_more}
\end{table*}

\begin{table*}[t!]
\setlength{\tabcolsep}{3.5pt}
\aboverulesep=0ex
\belowrulesep=0ex 
\centering
\footnotesize
\begin{tabular}{lrccccc}
\toprule
\multirow{2}{*}{\textbf{Mask-Referred Regional Description} (VideoRefer-Bench\textsuperscript{D}~\cite{yuan2025videorefer})} & \makecell{Samples \\ Added (\%)} & \multirow{2}{*}{\textbf{Avg.}} & \makecell{Subject \\ Correspondence} & \makecell{Temporal \\ Description} & \makecell{Appearance \\ Description} & \makecell{Hallucination \\ Detection} \\
\midrule
Baseline Ablation Model
& N/A   & $2.7308$ & $3.5200$ & $2.4235$ & $2.5639$ & $2.4160$ \\
Baseline Ablation Model + SoM~\cite{yang2023set} & N/A   & $2.6593$ & $3.5600$ & $2.1834$ & $2.3576$ & $2.5363$ \\
\bottomrule
\end{tabular}
\caption{\textbf{Regional Description} results on VideoRefer-Bench\textsuperscript{D} before and after applying the visual prompting method, SoM~\cite{yang2023set}. 
}
\label{tab:res_region_cap_som}
\end{table*}

\begin{table*}[t!]
\setlength{\tabcolsep}{2.8pt}
\small
  \aboverulesep=0ex
  \belowrulesep=0ex 
\centering
\begin{tabular}{lr|c}
\toprule
Timestamp-Referred QA (Yes/No) & Samples Added (\%) & QVHighlights~\cite{qvhighlights} \\
\midrule
Baseline Ablation: Video Instruction-Tuning Data~\cite{blip3video}  & N/A  & $0.5297$ \\
Baseline Ablation: Video Instruction-Tuning Data~\cite{blip3video} + NumberIt~\cite{wu2025number}  & N/A  & $0.5301$ \\
\rowcolor{LightGray!15}  
Baseline: Base Recipe (1,948,679 samples) & N/A    & $0.5288$ \\
\rowcolor{LightGray!15}  Baseline: Base Recipe (1,948,679 samples) + NumberIt~\cite{wu2025number}  & N/A    & $0.5311$ \\ \midrule
\rowcolor{LightSkyBlue!20} \quad   + $\mathcal{G}6$ + $\mathcal{G}7$  + $\mathcal{G}8$  + Remaining $\mathcal{G}5$ (Sampled) + $\mathcal{G}1$ & $28.73\%$  & $\mathbf{0.6031}$  \\
\rowcolor{LightSkyBlue!20} \quad   + $\mathcal{G}6$ + $\mathcal{G}7$  + $\mathcal{G}8$  + Remaining $\mathcal{G}5$ (Sampled) + $\mathcal{G}1$ + NumberIt~\cite{wu2025number}  & $28.73\%$   & $\mathbf{0.6041}$  \\
\bottomrule
\end{tabular}
\caption{\textbf{Timestamp-based Yes/No QA} results on QVHighlights before and after applying the visual prompting method, NumberIt~\cite{wu2025number}. 
} 
\label{tab:res_numberIt}
\end{table*}

\begin{table*}[t]
\centering
\begin{tabular}{c}
\toprule
Prompt = ``child in a white T-shirt.child in a pink top.dog with a red leash.woman with ponytail.'' \\
\midrule
\textbf{Video Frames} \\
\includegraphics[width=0.9\linewidth]{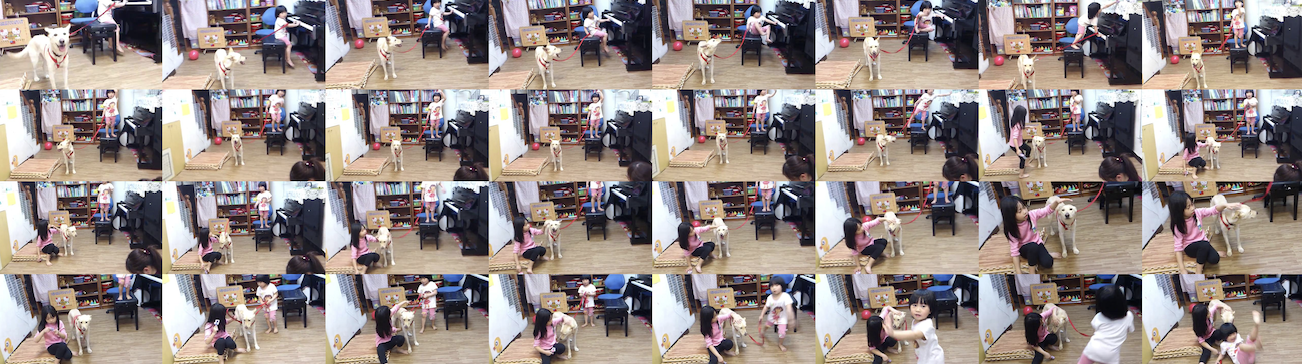}  \\
\midrule
\textbf{Ours} \\
\includegraphics[width=0.9\linewidth]{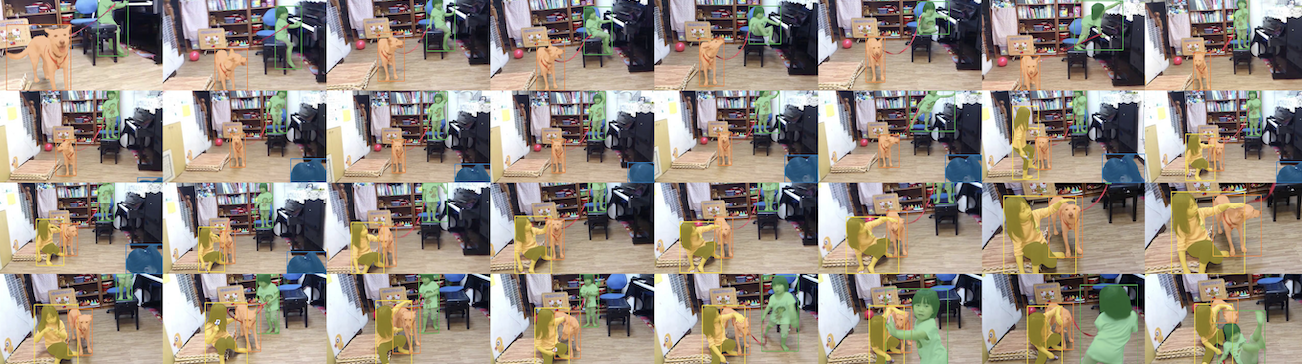}  \\
\colorbox{highlight_g}{child in a white T-shirt}~\colorbox{highlight_y}{child in a pink top}~\colorbox{highlight_o}{dog with a red leash}~\colorbox{highlight_b}{woman with ponytail} \\
\midrule
\textbf{Prior Method - GroundedSAM2~\cite{ren2025grounded_sam2}} \\
\includegraphics[width=0.9\linewidth]{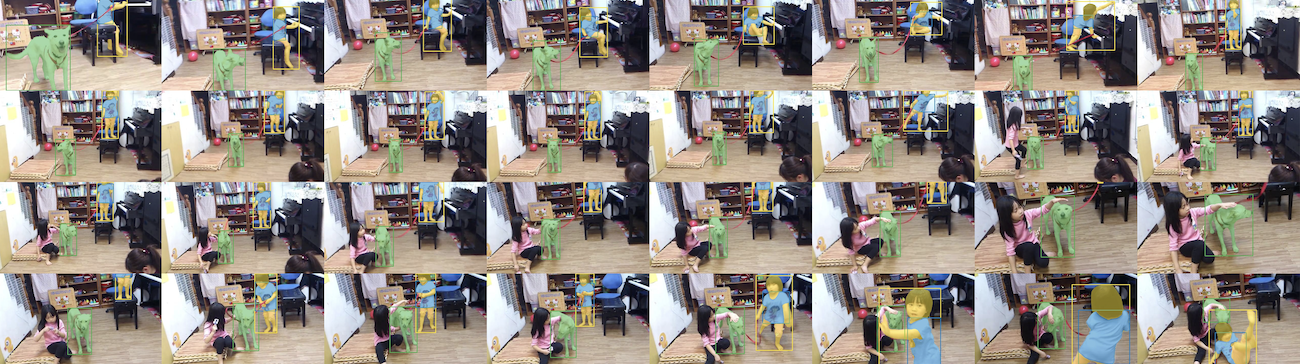}  \\
\colorbox{highlight_b}{child}~\colorbox{highlight_g}{dog}~\colorbox{highlight_y}{child woman} \\
\bottomrule
\end{tabular}
\captionof{figure}{\textbf{Qualitative Results of Referring Masklet Generation.} In this video, our method accurately generates masklets corresponding to the input referring expressions. In contrast, GroundedSAM2~\cite{ren2025grounded_sam2} fails to assign expressions to masklets and does not detect the woman and the child in a pink top, who appear midway through the video.}
\label{tab:quali_r2m_2}
\end{table*}

\begin{table*}[t]
\centering
\begin{tabular}{c}
\toprule
Prompt = ``woman on a bicycle.man in a blue shirt.man in a white shirt.man in a black shirt.'' \\
\midrule 
\textbf{Video Frames} \\
\includegraphics[width=0.9\linewidth]{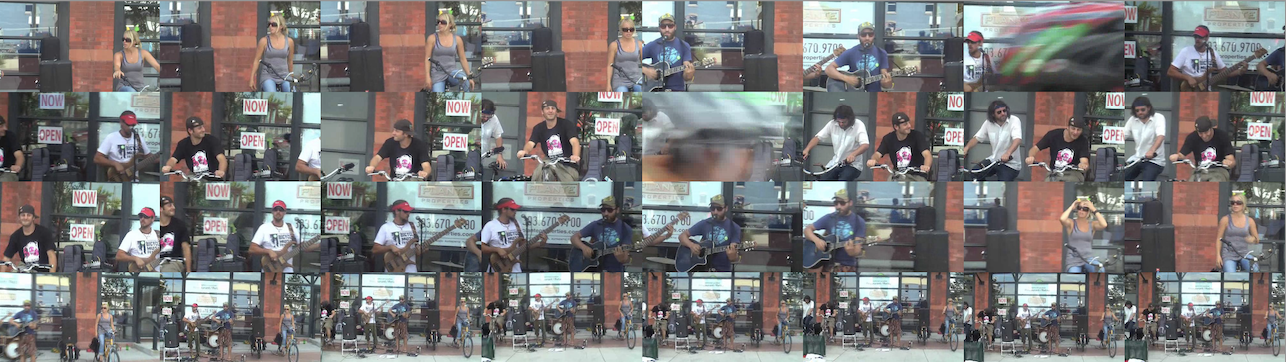}  \\
\midrule
\textbf{Ours} \\
\includegraphics[width=0.9\linewidth]{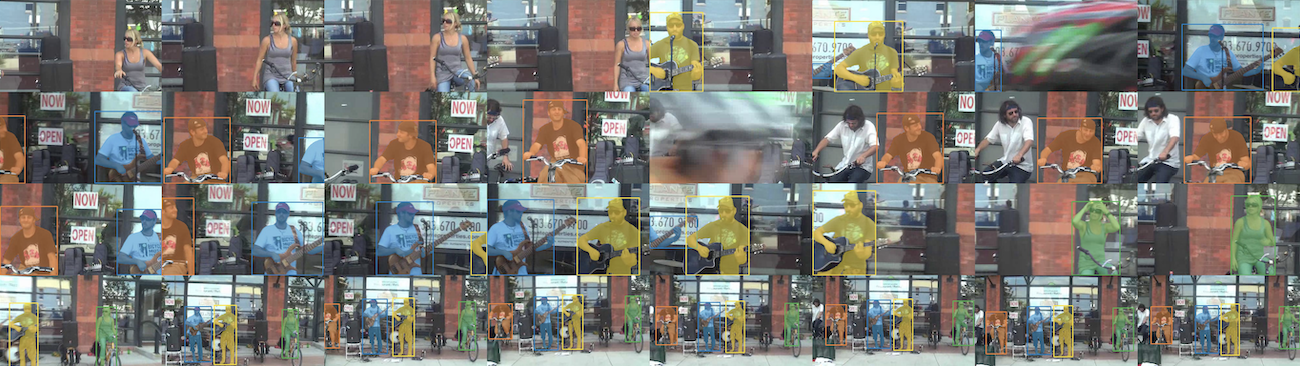}  \\
\colorbox{highlight_g}{woman on a bicycle}~\colorbox{highlight_y}{man in a blue shirt}~\colorbox{highlight_o}{man in a black shirt}~\colorbox{highlight_b}{man in a white shirt} \\
\midrule
\textbf{Prior Method - GroundedSAM2~\cite{ren2025grounded_sam2}} \\
\includegraphics[width=0.9\linewidth]{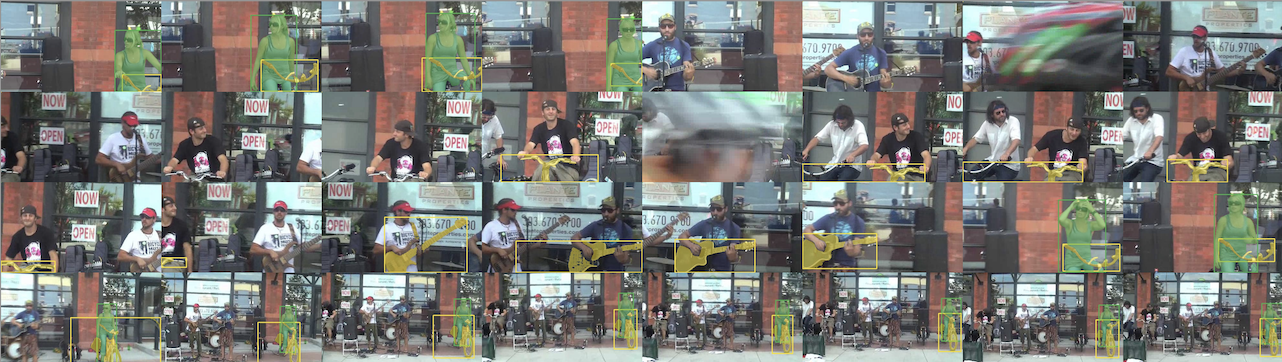}  \\
\colorbox{highlight_g}{woman}~\colorbox{highlight_y}{bicycle} \\
\bottomrule
\end{tabular}
\captionof{figure}{\textbf{Failure Results of Referring Masklet Generation.} Our method fails to consistently track the woman on a bicycle throughout the video, while GroundedSAM2~\cite{ren2025grounded_sam2} fails to detect, track, and differentiate the individuals referenced in the input text prompt. Videos with heavy motion blur and long-range dependencies remain challenging to handle.}
\label{tab:fquali_r2m_1}
\end{table*}

\begin{figure*}[t]
	\centering
	\vspace{-10pt}
    \includegraphics[width=1.0\linewidth]{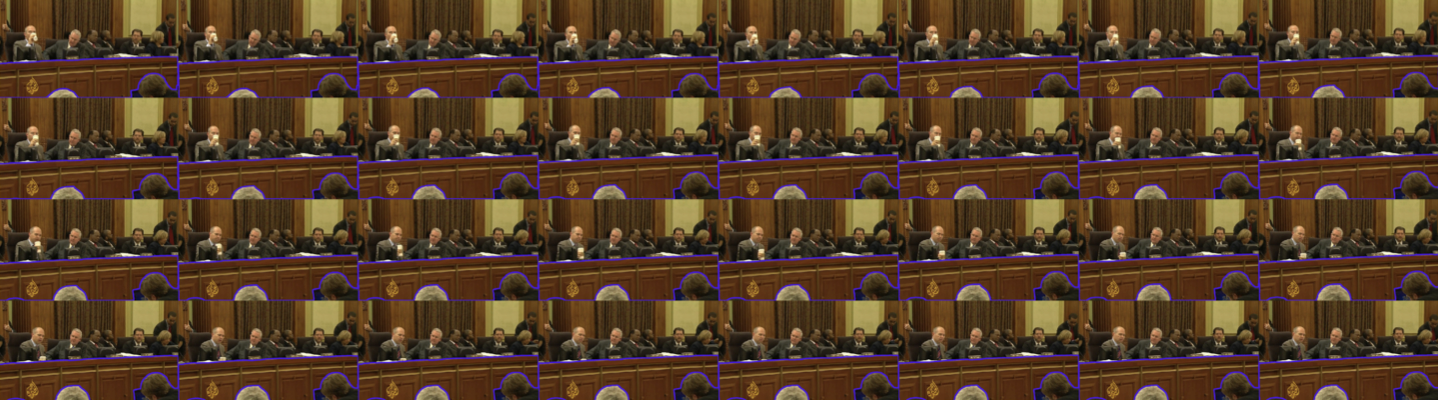}
    \includegraphics[scale=0.7]{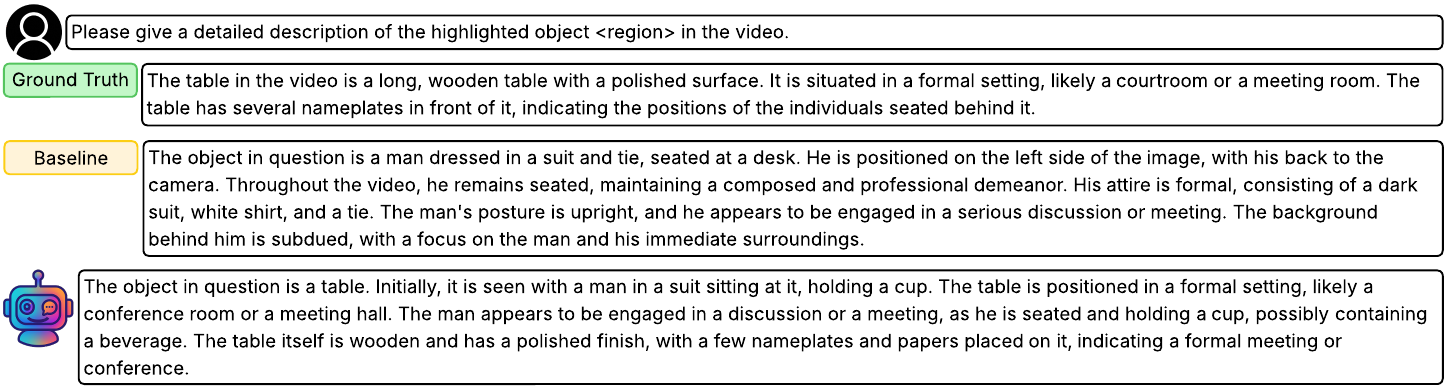}
    \vspace{-1pt}
\captionof{figure}{\textbf{Qualitative Results of Video LLMs Trained w/ and w/o \ourseos-Synthethized Data}. 
This sample is drawn from VideoRefer-Bench\textsuperscript{D}, designed to assess a model's performance on the task of \textbf{Mask-Referred Regional Description}. 
The boundary of the region referred to by the mask in this sample is highlighted in purple. While the video includes several individuals as prominent foreground elements, the masklet specifically refers to the table, not the people. The baseline model, however, fails to interpret the mask correctly and mistakenly answers that the referred object is a man. In contrast, the model trained on \ourseos-generated data accurately identifies the masklet-referred region as a table.
}
    \label{fig:quali_good_benchd_table}
    \vspace{-10pt}
\end{figure*}

\begin{figure*}[t]
	\centering
    \includegraphics[width=1.0\linewidth]{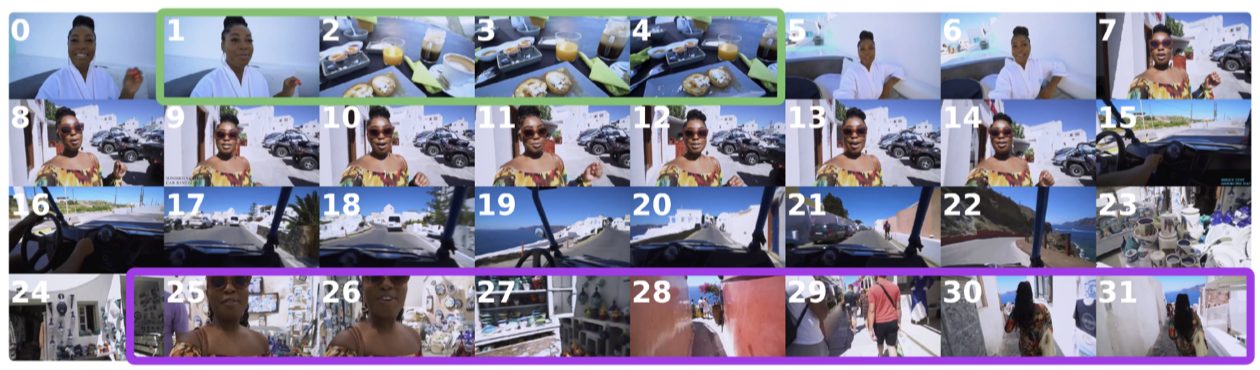}
    \includegraphics[scale=0.70]{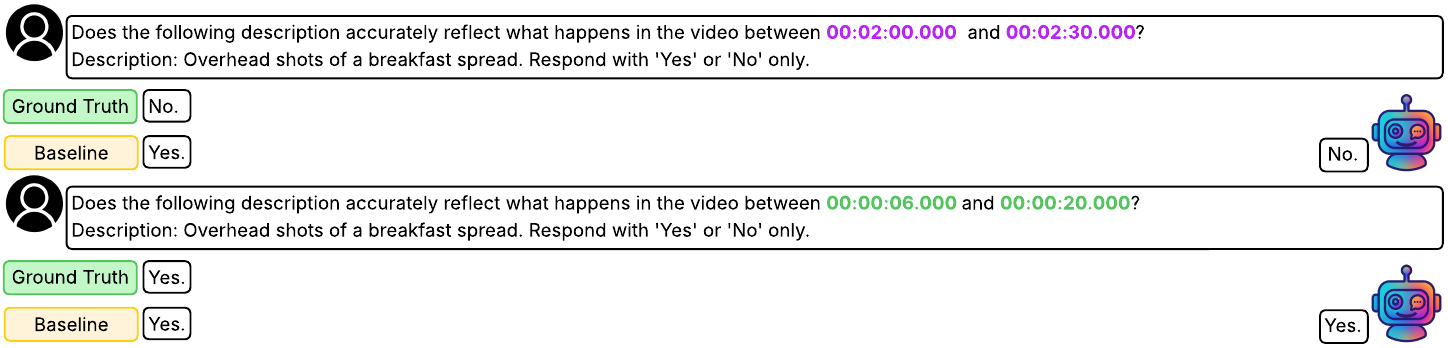}
\captionof{figure}{\textbf{Qualitative Results of Video LLMs Trained w/ and w/o \ourseos-Synthethized Data}. 
This sample is drawn from QVHighlights, using our repurposed task designed to assess a model's performance on \textbf{Timestamp-Referred Video QA}. 
The segment boundaries corresponding to the timestamps in the first and second questions are highlighted in purple and green, respectively. 
The model trained on our \ourseos-generated data correctly answers both questions, demonstrating superior understanding of precise moments and segments in videos compared to the baseline. 
}
    \label{fig:quali_good_qv_1}
\end{figure*}

\subsubsection{Discussions and Insights}
Using \ourseos, we generated $947,854$ instruction data points from just $4,253$ NExT-QA videos. While our \colorbox{LightSkyBlue!20}{final recipe} added only $545$ more videos compared to the baseline, it led to performance gains across multiple benchmarks with both short and long videos.

This highlights a key insight: the value of video instruction-tuning data is not merely in its quantity, but in its quality and specificity. \textbf{Rather than indiscriminately scaling the dataset with more videos and generic questions, we found that carefully crafting meaningful, well-grounded questions and answers leads to significantly better training outcomes for video-language models}.

We identify the following three critical principles for curating effective video instruction data:

\begin{itemize}
     \item \textbf{Video-\textit{grounded} instruction-response pairs:} Both questions and responses must be tightly linked to the video content. \ours achieves this through spatiotemporally grounded metadata generated using multiple pixel-level vision foundation models.
     
    \item \textbf{Temporal and fine-grained reasoning emphasis:} Effective data should challenge models to focus on dynamics and reason about time, especially those involving fine-grained details and long-range dependencies in both space and time.
    
    \item \textbf{Diverse tasks and formats:} A mix of 
    task types and instruction data formats
    ensures broader coverage and more robust video understanding and reasoning capabilities.
\end{itemize}

\subsubsection{Visual Prompting for Space-Time Referring}
\label{subsubsec:training_free}

It is worth noting that incorporating the plug-and-play modules and modify the architecture of the pre-trained general-purpose Video LLM for space-time referring is not strictly necessary. 
We explore visual prompting approaches: SoM~\cite{yang2023set} for masklet comprehension and NumberIt~\cite{wu2025number} for timestamp understanding.

\noindent \textbf{SoM: Mask-Overlay-Frame Prompting}. We follow the implementation of the Set-of-Mark (SoM) method from VideoRefer~\cite{videorefer_eval_2024} to apply masks to video frames, as originally proposed by~\cite{yang2023set}. We also changed the question prompt into: ``\texttt{I have outlined an object with a red contour in the video. Please describe the object in detail.}''
The results are presented in Table~\ref{tab:res_region_cap_som}. After applying SoM, the average performance on the Mask-Referred Video Regional Description task decreases, but performance increases on certain metrics, e.g., Subject Correspondence. 

Our analysis reveals that the effectiveness of SoM is highly sensitive to the way masks are rendered on the video frames. In our initial implementation, we used thicker mask boundaries and semi-transparent red fill color. This approach led to severe hallucinations by the model, which often misinterpreted masked regions as merely red-colored objects. In contrast, the SoM implementation from VideoRefer~\cite{videorefer_eval_2024} uses thinner boundaries and fully transparent fills, resulting in significantly improved performance over our version. Nevertheless, the performance remains lower than the baseline without any SoM prompting.

\noindent \textbf{NumberIt: FrameID-Overlay-Frame Prompting}. Similar to SoM, NumberIt~\cite{wu2025number} overlays the frame ID at a specific location on each frame. We overlaid the frame ID in red, following the authors' suggestion, and placed each ID in the top-left corner of the corresponding frame (the resulting rendering effect is similar to Fig.~\ref{fig:quali_good_qv_2}). We also modified each question as follows:
\texttt{The red numbers on each frame represent the frame number. Does the following description accurately reflect what happens in the video between <frame\_start> and <frame\_end>?\\Description: \{description\}. Respond with `Yes' or `No' only.} For each question, we substituted \texttt{<frame\_start>} and \texttt{<frame\_end>} with their corresponding frame IDs.
The results are listed in Table~\ref{tab:res_numberIt}.
The performance on timestamp-based Yes/No Video QA in QVHighlights shows a slight improvement after applying the visual prompting method, NumberIt.

\textbf{Summary}: Using a pretrained general-purpose Video LLM for space-time referring tasks does not necessarily require altering the model architecture.
We found that visual prompting approaches, such as SoM and NumberIt, can help the model perform mask-referring or timestamp-referring tasks; however, their effectiveness appears limited in the absence of model tuning.
We hypothesize that incorporating these techniques during model fine-tuning—while preserving the original architecture—may lead to more performance gains~\cite{wu2025number}, which we leave for future work.

\subsection{Qualitative Results}
\label{subsec:quali_results}

In this section, we present qualitative results that best illustrate the strengths and limitations of our method and model. We discuss limitations in detail in the next section.

\subsubsection{\ourseos-Synthesized Instruction Data}

We present qualitative results of our \ourseos-synthesized data
in Fig.~\ref{fig:quali_good_ourdata_2} and Fig.~\ref{fig:quali_good_ourdata_1} in the Appendix.
Despite some 
noise 
in the pseudo-annotated video metadata, the synthesized instruction-response pairs capture essential spatio-temporal dynamics from the videos and successfully enhance the model's performance.

\subsubsection{Our Referring Masklet Generation}

We present qualitative results of our novel referring masklet generation pipeline in Fig.~\ref{tab:quali_r2m_2}, 
Fig.~\ref{tab:fquali_r2m_1},
Fig.~\ref{tab:quali_r2m_1} and Fig.~\ref{tab:quali_r2m_3} (some figures are in the Appendix). 
Our referring masklet generation pipeline demonstrates superior performance compared to the widely adopted prior method, GroundedSAM2~\cite{ren2025grounded_sam2}.

\subsubsection{\ourseos-Trained Model}

We also present comprehensive qualitative results of our \ourseos-trained model in comparison to the `Baseline' model.
We observe that the models tend to exhibit a foreground bias and misinterprets the masklet. The model trained on our data mitigates this limitation (see Fig.~\ref{fig:quali_good_benchd_table}, Fig.~\ref{fig:quali_good_benchd_parking}). In cases requiring entity disambiguation (e.g., Fig.~\ref{fig:quali_good_benchq_panda}, Fig.~\ref{fig:quali_good_benchq_cat}, Fig.~\ref{fig:quali_good_benchq_twoplayers}, and Fig.~\ref{fig:quali_good_benchq_twomonkeys}), our model produces more accurate results. These gains stem from two design choices: \textbf{(1)} our LLM-based question generation strategy ensures that questions explicitly reference masklet information, and \textbf{(2)} our carefully designed masklet generator addresses the limitations of GroundedSAM2 in handling such scenarios (in contrast, GroundedSAM2 is used by VideoRefer-700K~\cite{yuan2025videorefer} in its data generation process). Additionally, our model demonstrates superior fine-grained spatiotemporal action understanding (see Fig.~\ref{fig:quali_good_benchq_bike}) and exhibits superior performance in precise timestamp-based video comprehension (i.e., Fig.~\ref{fig:quali_good_qv_1}, Fig.~\ref{fig:quali_good_qv_2}, and Fig.~\ref{fig:quali_good_qv_3}).

\section{Limitations and Future Directions}
\label{sec:limit_future}

\ours synthesized data is not error-free.
For example, in the last blue-highlighted video segment shown in Fig.~\ref{fig:quali_good_ourdata_1} of the Appendix, \ours identifies the woman as not present. However, a human viewer would easily identify the woman in that segment while watching the video, despite the fact that she is largely occluded and the frames are mostly occupied by the child. This error also occurs because we did not employ a more complex, dense video captioning framework that leverages \textit{inter-segment} information, such as a hierarchical~\cite{zhang2024videoinstructiontuningsynthetic} or differential video captioning~\cite{chen2024sharegpt4video} method. We actually tried these approaches, but they did not yield better results than clip-by-clip captioning using current open-source models. 
We also experimented with several alternative open-source mid-scale Video LLMs, including Qwen2.5-VL-7B-Instruct, LLaVA-NeXT-Video-34B, LLaVA-OneVision-7B, and Tarsier-7B. Ultimately, we selected Tarsier-34B, as it appeared to provide more accurate, action-centric descriptions.

Scenes characterized by high visual clutter and significant dynamic variations continue to pose substantial challenges. 
Fig.~\ref{tab:fquali_r2m_1} illustrates a failure case of our referring masklet generation—the woman is not tracked or segmented in the first four frames. This highlights that videos with heavy motion blur and long-range dependencies remain challenging to handle, even for our method. The issue stems from the tracking limitations of SAM2, the tracking and segmentation model we employ which is sensitive to the selection of the start tracking frame. Fig.~\ref{fig:quali_good_ourdata_2} presents another example of  tracking and segmentation failure—the child in the black shirt lacks associated masks in frame $10$ and frame $11$, despite being clearly visible.

\ours may inherit limitations from the underlying models used in its modular components, such as pixel-level foundation models, LLMs, and Video LLMs. For instance, LLMs and Video LLMs are known to hallucinate, potentially introducing misleading information into annotated video metadata and synthesized question–answer pairs. Similarly, the extraction of pixel-level information may be less reliable in videos with highly similar individuals or densely populated scenes (e.g., crowded urban environments), which the pixel-level foundation models we used could not reliably handle. Despite these challenges, the modular nature of \ours positions it well to benefit from future improvements in its underlying models—including LLMs, Video LLMs, and grounding vision foundation models—as they continue to evolve.

\ours involves multiple models, which may hinder exact reproduction of its pseudo-annotation, data synthesis as well as our model training procedures for research groups with limited resources. However, \ours is a multi-stage modular framework, wherein individual model components can be substituted with more computationally efficient alternatives, enabling flexible adaptation under varying resource constraints.

In terms of the limitations inherent to models trained on \ourseos-synthesized data, Fig.~\ref{fig:quali_failure_1} shows a typical failure case where the region indicated by the mask is neither the primary foreground object nor centrally positioned in the frame. This reflects a common issue across all models, including baselines, which tend to exhibit both a foreground main object bias and a center bias. Furthermore, Fig.~\ref{fig:quali_failure_2} illustrates another failure case, underscoring the persistent difficulty in capturing fine-grained action semantics.

Future work may further improve \ours by refining individual modules—for example, improving the video clipping pipeline to produce entity-centric segments, and incorporating feedback-verification mechanisms to minimize hallucinated content in video metadata and instruction-following pairs. Additionally, given the potential for error propagation in our modular framework, as well as the nature of synthetic data, which may not fully match real human question distributions, future research is encouraged to develop effective filtering strategies for the synthetic instruction-tuning data as a final quality assurance or adjustment step in the data engine, thereby enabling more efficient, reliable, and robust model training.

For the development of the referring and reasoning video model, our current trained models are limited to mask-based spatial referring and is not trained for other types of spatial references such as points, boxes, and scribbles. However, since these other forms of spatial references are inherently sparser than masks and can be easily derived from object segmentation masks, an avenue for future research is to explore transforming the current mask-level instruction-tuning data produced by \ours into alternative data formats, and to train models that can comprehend more diverse forms of user spatial references.

Moreover, the LLM backbone of models we trained on our data is based on the pretrained microsoft/Phi-3-mini-4k-instruct~\cite{abdin2024phi}. As with many other Video LLMs, the performance of our model is heavily influenced by the capabilities of the underlying LLM. We encourage further research into training Video LLMs on \ourseos-synthesized data using larger and more powerful models. However, this typically demands significantly more tuning data and greater computational resources.

In our experiments, we conducted rather limited exploration of the optimal training data mixture. As a result, the current composition may not represent the most effective setup for fostering broad, balanced and transferable skills. Future work could focus on systematically optimizing the data composition, which is likely to result in more substantial and consistent performance gains across diverse benchmarks and metrics.

Finally, our model is grounded at the perception level rather than at the output response generation level. While grounding at the output level offers a more direct path to interpretable video-language reasoning, it requires training data with high-fidelity spatiotemporal annotations. At present, the boundary of the fine-grained space-time information generated by \ours may lack the precision required to reliably supervise such models. 

Our work centers on \textit{referring} understanding-where the model leverages fine-grained spatiotemporal cues as conditional input, and our models are trained using synthesized instruction-tuning data, rather than being directly supervised by pseudo-annotated dense video metadata. This setup is inherently more robust to moderate imperfections—such as missing entities or imprecise temporal boundaries. For example, as shown in Fig.~\ref{fig:quali_good_ourdata_2}, even when the temporal span of the masklet for the child in the black shirt is incomplete, the associated instruction-response pair remains accurate and meaningful. This resilience arises because referring understanding does not require an exhaustive coverage of the language-described pixel-level space-time information, making it more adaptable under the current data limitations.

Looking ahead, advancing output-level spatiotemporal grounding in Video LLMs holds significant promise for improving their generalization, reliability and fine-grained spatiotemporal reasoning skills. We encourage future work to pursue this direction by leveraging more accurate spatial-temporal annotations aligned with language, ideally enabled by an enhanced, scalable, and automated data generation pipeline.

\section{Related Work}
\vspace{-5pt}

\vspace{5pt}
\noindent \textbf{Video Spatial Referring}.
A growing body of research has enabled Multimodal Large Language Models (MLLMs) to perform regional understanding in static images~\cite{yan2024list,zhang2025gpt4roi,yuan2024osprey,ma2024groma,rasheed2024glamm,xu2024pixel,zhang2024ferret,you2023ferret,zhang2024llava}, but
spatial referring understanding
in videos remains underexplored. 
Earlier efforts include Artemis~\cite{qiu2024artemis} for single-object referencing using 
box-level representations, as well as Elysium~\cite{wang2024elysium} and Merlin~\cite{yu2024merlin} that transform 
box coordinates 
into textual prompts to help the LLM identify the referred objects.
Nevertheless, these methods are plagued by imprecise regional understanding.

Recently, a new wave of concurrent work
has shifted towards mask-based region-level understanding in videos.
DAM~\cite{lian2025describe} and PAM~\cite{lin2025perceive}, while demonstrating extension to videos, primarily focus on regional captioning rather than general instruction-following.
Both Omni-RGPT~\cite{heo2025omni} and SAMA~\cite{sun2025sama} introduced automatic annotation methods using GPT-4o and Gemini 1.5 Pro, resp., to transform existing video datasets with annotated regions into conversational data. VideoRefer~\cite{yuan2025videorefer} introduces a multi-agent annotation pipeline that operates without the need for pre-existing annotations. However, its design struggles with complex videos such as scenes involving multiple entities of the same category or cases where entities are temporarily absent. Additionally, it cannot produce temporal annotations, 
limiting its ability to support time-sensitive tasks.

\vspace{5pt}
\noindent \textbf{Video Timestamp Referring}.
We posit that enabling models to comprehend timestamp-specific video segments
may enhance their video temporal reasoning skills. Therefore, we propose \ourseos, a framework that generates timestamp-based temporal annotations and instructions for videos to empower model to understand video timestamp-based content referencing. To the best of our knowledge, there is no existing work that focuses solely on this task. However, there has been a line of research on temporally grounded Video LLMs~\cite{qian2024momentor,wang2024grounded,wu2025number,guo2025vtg,zhao2025videoexpert,huang2023vtimellm}, which are capable of both understanding and localizing temporal boundaries
in response to user queries. These models
require training on 
boundary-accurate 
temporal localization data,
and we consider temporal or spatiotemporal localization a future work.

\vspace{5pt}
\noindent \textbf{Video Instruction-Following Data}.
Early efforts such as VideoChat~\cite{li2023videochat}, Video-ChatGPT~\cite{maaz2023video} and MVBench~\cite{li2024mvbench} convert existing video annotations into a conversational format. Later on, more studies employed proprietary models to produce training data
such as ShareGPT4Video~\cite{chen2024sharegpt4video} and MiraData~\cite{ju2024miradata} for the video captioning task, but obtaining versatile and effective video instruction-following data has remained challenging, with earlier datasets criticized for their limited utility~\cite{zhang2024videoinstructiontuningsynthetic}. LLaVA-Video-178K~\cite{zhang2024videoinstructiontuningsynthetic} then takes a step forward by sourcing existing video datasets (with videos up to $3$ minutes) and enriching them with open-ended and multiple-choice questions across diverse tasks, developed through a combination of GPT-4o and human efforts. LongViTU~\cite{wu2025longvitu} and VideoMarathon~\cite{lin2025unleashing} were also introduced for hour-scale long video training. 

Inspired by these efforts—as well as the TempCompass benchmark~\cite{liu2024tempcompass}, which emphasizes the diversity of temporal aspects and task formats, we design instruction-following tasks and formats that capture a wide range of temporal variations using pseudo-annotated video metadata. 
We are also influenced by T3~\cite{li2025videot3}, which shows that with proper temporally focused task design, textual temporal instruction data can enhance temporal reasoning without video-based training. However, we argue that true spatiotemporal reasoning of an AI agent requires the video input. Our synthetic instruction data target space-time referring understanding that encourages perceptually-grounded, instruction-tuned Video LLMs.

\vspace{5pt}
\noindent \textbf{Positioning of Our Work}.
None of the above general video instruction data is designed to empower Video LLMs for the fine-grained, space-time referring tasks. Currently, the only available instruction tuning data for masklet referring comprehension in videos is VideoRefer-700K~\cite{yuan2025videorefer}, which we added into our base training recipe but it lacks temporally focused tasks and annotations. Unlike prior work, \ours pseudo-annotates spatiotemporally dense metadata—subjects, objects, their locations as masklets, as well as their action descriptions and 
timelines—for complex video scenarios. It generates spatially and temporally fine-grained, grounded instruction-response pairs, requiring no legacy annotations. Without proprietary models or the need to annotate large volumes of new videos, instruction data from \ours empowers models for space-time referring and spatiotemporal reasoning.

\section{Conclusion}

In this work, we introduced \ourseos, a novel synthetic instruction data generation framework designed to equip Video LLMs with fine-grained, space-time referring and reasoning
capabilities. By leveraging temporally dense, pseudo-annotated video metadata, \ours systematically produces instruction-response pairs that are richly grounded in space and time. This enables models to better resolve complex user queries involving object-centric events, temporal references, and gestural or spatial cues—challenges that existing Video LLMs struggle to address.

Our experiments demonstrate that models trained with \ours data outperform baselines on tasks requiring spatiotemporal disambiguation, confirming the efficacy of our approach in fostering more perceptually grounded and context-aware video understanding. \ours thus represents a foundational step toward building next-generation AI companions capable of realistic human-AI interactions and sophisticated spatiotemporal reasoning. 
Future work may explore effective methods for filtering synthetic data and advancing the output-level spatiotemporal grounding capabilities of Video LLMs, further enhancing their generalization and reasoning performance.

{
    \small
    \bibliographystyle{ieeenat_fullname}
    \bibliography{main}
}

\clearpage

\clearpage

\newpage
\appendix

\section{Appendix / Supplemental Material}

\noindent \hyperref[app_sec:quali]{\textcolor{black}{\textbf{A.1. More Qualitative Results}}}

\quad \hyperref[app_subsec:quali_strefer]{\textcolor{black}{\textbf{A.1.2 \ourseos-Synthesized Data}}}

\quad \hyperref[app_subsec:quali_model]{\textcolor{black}{\textbf{A.1.2 \ourseos-Trained Model}}}

\noindent \hyperref[app_sec:strefer_details]{\textcolor{black}{\textbf{A.2. More \ours Details}}}

\noindent \hyperref[app_sec:model_details]{\textcolor{black}{\textbf{A.3. Model Details}}}

\quad \hyperref[app_subsec:arch_overview]{\textcolor{black}{\textbf{A.3.1 Architecture Overview}}}

\quad \hyperref[app_subsec:video_token]{\textcolor{black}{\textbf{A.3.2 Video Token Representation}}}

\quad \hyperref[app_subsec:mask_token]{\textcolor{black}{\textbf{A.3.3 Masklet Reference Token Representation}}}

\quad \hyperref[app_subsec:timestamp_token]{\textcolor{black}{\textbf{A.3.4 Timestamp Reference Token Representation}}}

\bigskip
\bigskip

\subsection{Additional Qualitative Results}
\label{app_sec:quali}
\subsubsection{Qualitative Results of \ourseos-Synthesized Data}
\label{app_subsec:quali_strefer}

\begin{figure*}[t]
	\centering
    \includegraphics[scale=0.57]{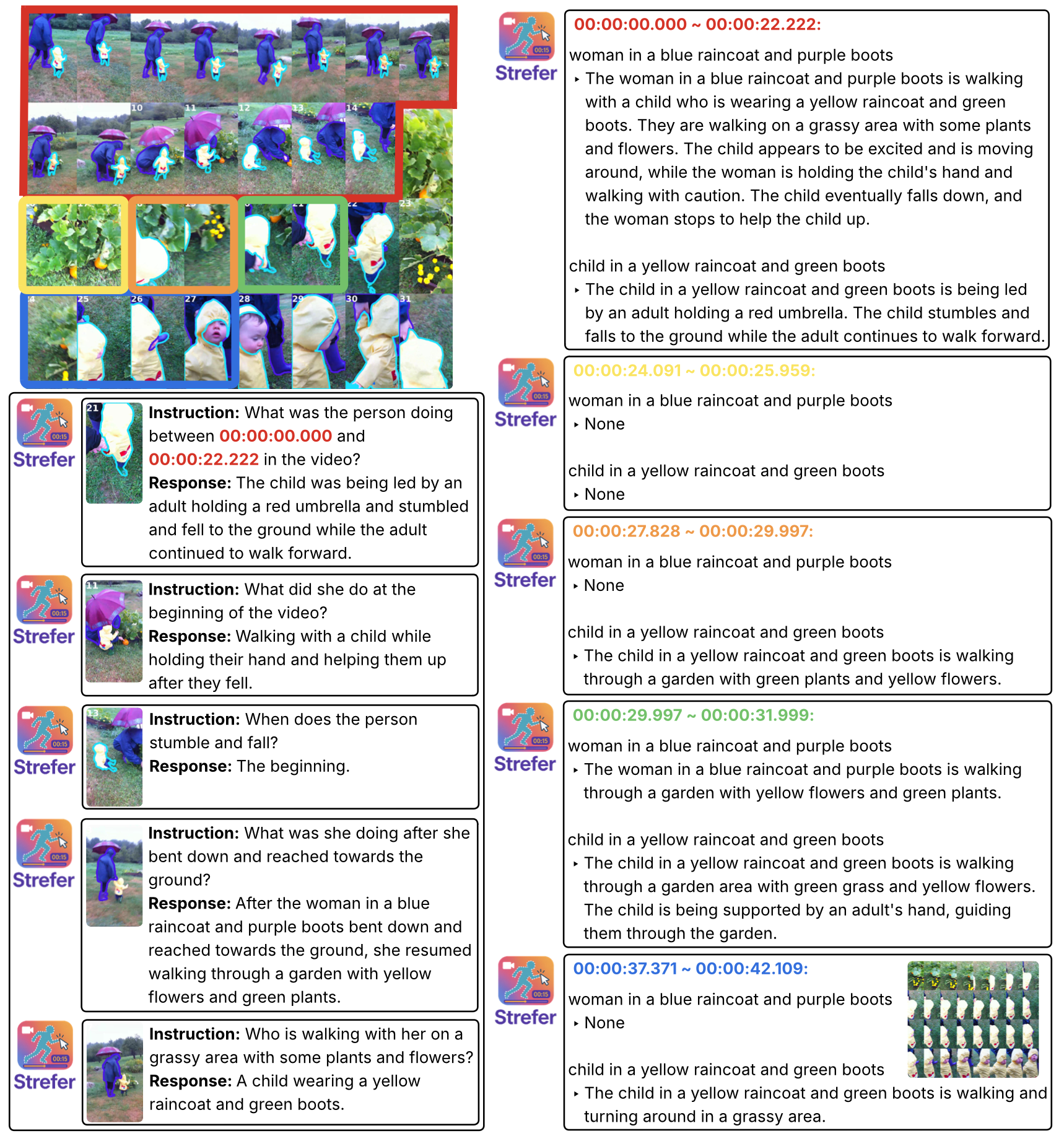}
    \vspace{-1pt}
\captionof{figure}{\textbf{Example of \ourseos-Synthesized Instruction-Response Pairs (left) and Pseudo-Annotated Video Metadata (right).}
Each instruction begins with the prefix: ``Please answer the following question about the $<$region$>$'' (and the prefix is omitted in the figure). For each instruction-response pair, the boundary of the object mask referred to by $<$region$>$ is shown next to the pair and highlighted in color.
\ours automatically clips the video into segments and pseudo-annotates the video metadata—including active entities, their locations (as masklets), and their action descriptions and timelines—for complex video scenarios, such as scenes containing multiple entities of the same category, and cases where entities do not appear in the first frame, or temporarily exit and re-enter the frame; based on the auto-generated video metadata, it produces instruction-response pairs,
requiring no legacy annotations or manual efforts. 
Though current implementation of \ours does not any use proprietary models, without the need to annotate large volumes of new videos, instruction data from \ours empowers models for space-time referring and spatiotemporal reasoning (\textit{ref.}~Table~\ref{tab:res_region_cap}, ~\ref{tab:res_region_qa}, and~\ref{tab:res_more}). 
}
    \label{fig:quali_good_ourdata_1}
\end{figure*}

We present qualitative results of our \ourseos-synthesized data
in Fig.~\ref{fig:quali_good_ourdata_2} and Fig.~\ref{fig:quali_good_ourdata_1}, as well as qualitative results of our novel referring masklet generation pipeline within \ours in Fig.~\ref{tab:quali_r2m_2}, Fig.~\ref{tab:fquali_r2m_1}, Fig.~\ref{tab:quali_r2m_1} and Fig.~\ref{tab:quali_r2m_3}. Detailed discussions of the qualitative results are presented in Sec.~\ref{subsec:quali_results} and Sec.~\ref{sec:limit_future} of the main paper.

\subsubsection{Qualitative Results of \ourseos-Trained Model}
\label{app_subsec:quali_model}

\begin{figure*}[t]
	\centering
    \includegraphics[width=1.0\linewidth]{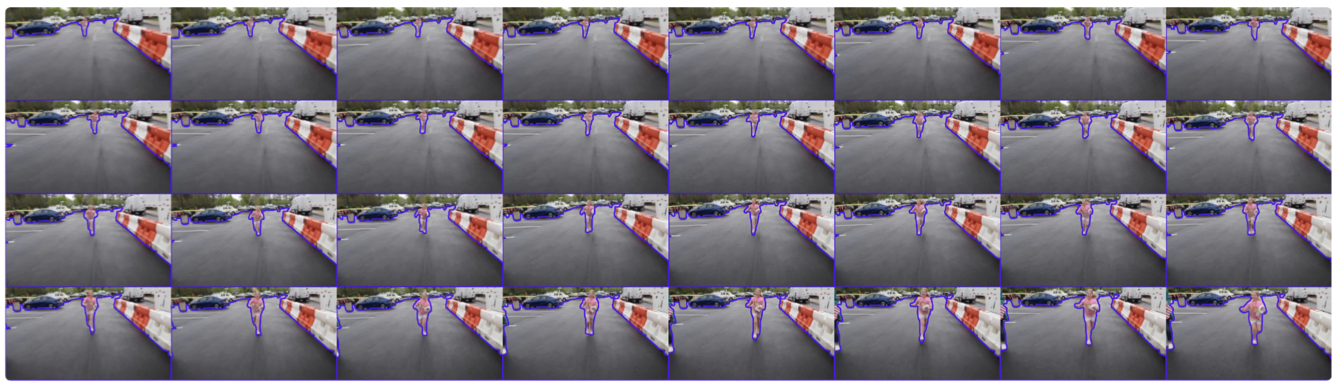}
    \includegraphics[scale=0.70]{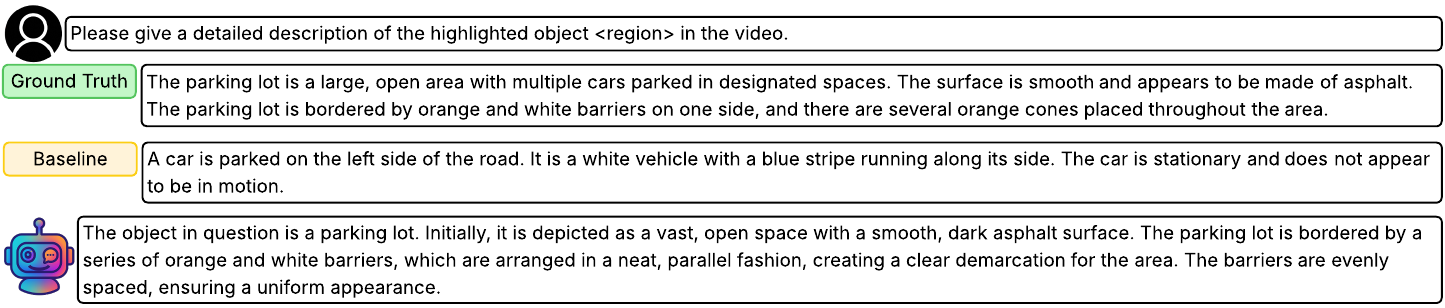}
    \vspace{-1pt}
\captionof{figure}{\textbf{Qualitative Results of Video LLMs Trained w/ and w/o \ourseos-Synthethized Data}. 
This sample is drawn from VideoRefer-Bench\textsuperscript{D}, designed to assess a model's performance on the task of \textbf{Mask-Referred Regional Description}. 
The boundary of the region referred to by the mask in this sample is highlighted in purple.
The baseline model tends to have a foreground bias, fails to interpret the masklet correctly, and mistakenly responds that the referred region is a car. In contrast, the model trained on \ourseos-generated data accurately identifies the masklet-referred region as a parking lot.
}
    \label{fig:quali_good_benchd_parking}
\end{figure*}

\begin{figure*}[t]
	\centering
    \includegraphics[width=1.0\linewidth]{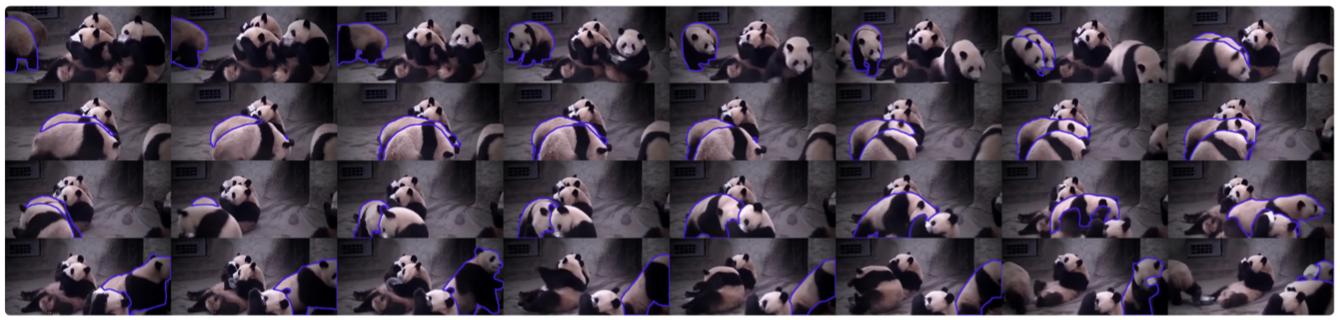}
    \includegraphics[scale=0.70]{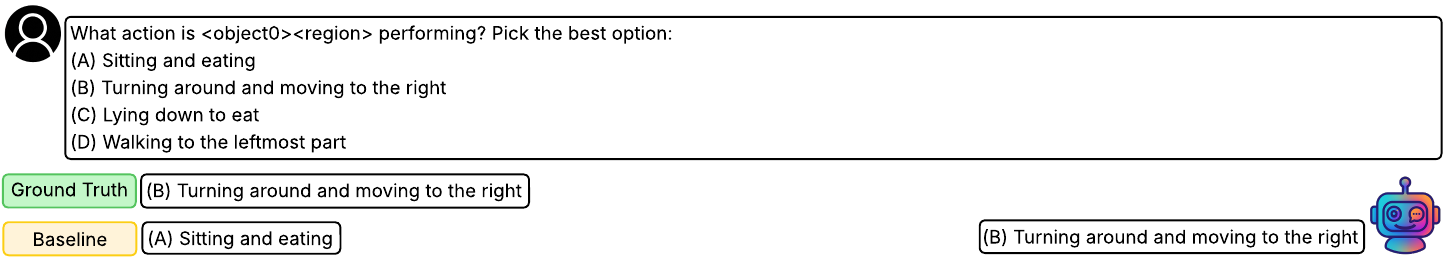}
\captionof{figure}{\textbf{Qualitative Results of Video LLMs Trained w/ and w/o \ourseos-Synthethized Data}. 
This sample is drawn from VideoRefer-Bench\textsuperscript{Q}, designed to assess a model's performance on the task of \textbf{Mask-Referred Regional QA}. 
The boundary of the region referred to by the mask in this sample is highlighted in purple.
The model trained on \ourseos-generated data correctly identifies the masklet-referred region and action.
}
    \label{fig:quali_good_benchq_panda}
\end{figure*}

\begin{figure*}[t]
	\centering
	\vspace{-5pt}
    \includegraphics[width=1.0\linewidth]{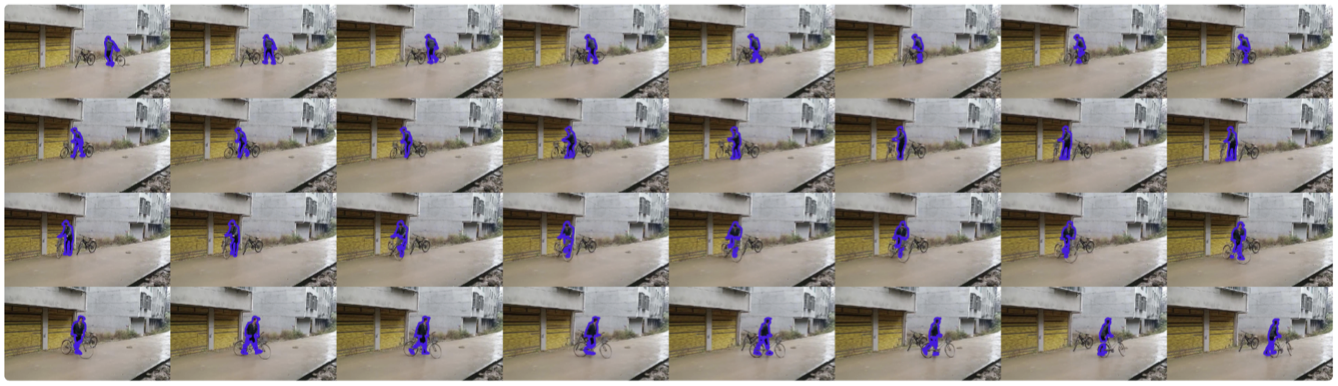}
    \includegraphics[scale=0.70]{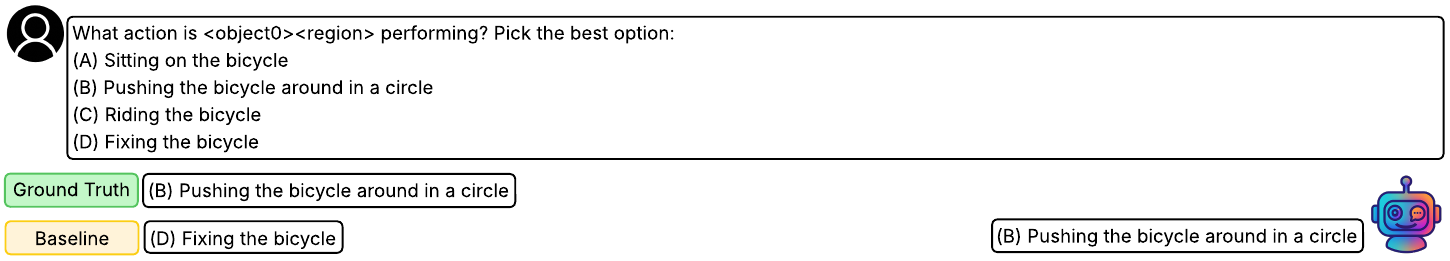}    
\captionof{figure}{\textbf{Qualitative Results of Video LLMs Trained w/ and w/o \ourseos-Synthethized Data}. 
This sample is drawn from VideoRefer-Bench\textsuperscript{Q}, designed to assess a model's performance on the task of \textbf{Mask-Referred Regional QA}. 
The boundary of the region referred to by the mask in this sample is highlighted in purple.
In this sample, the model
must demonstrate fine-grained spatiotemporal action understanding due to the small size of the mask and the subtle motion differences between the correct and negative options. The model trained on \ourseos-generated data successfully identifies both the region referred to by the masklet and the corresponding action.
}
    \label{fig:quali_good_benchq_bike}
\end{figure*}

\begin{figure*}[t]
	\centering
	\vspace{-5pt}
    \includegraphics[width=1.0\linewidth]{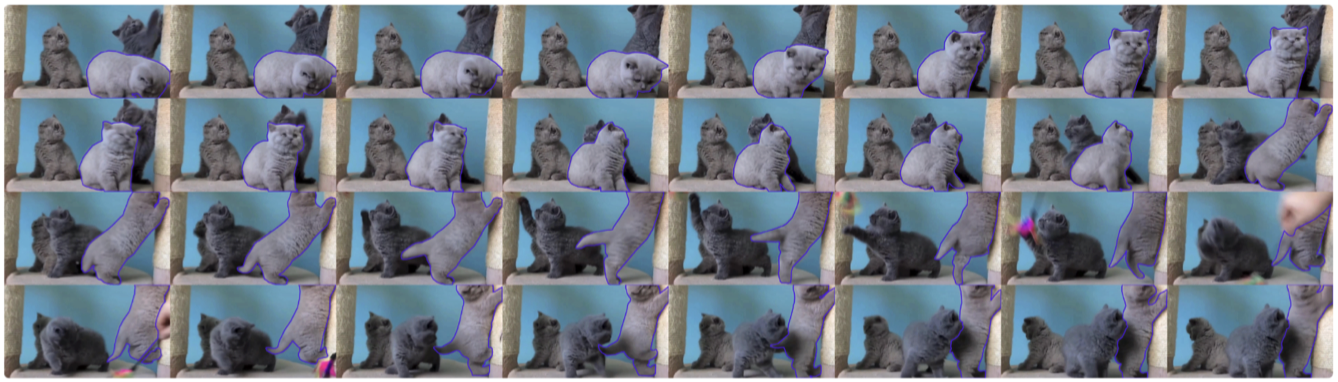}
    \includegraphics[scale=0.70]{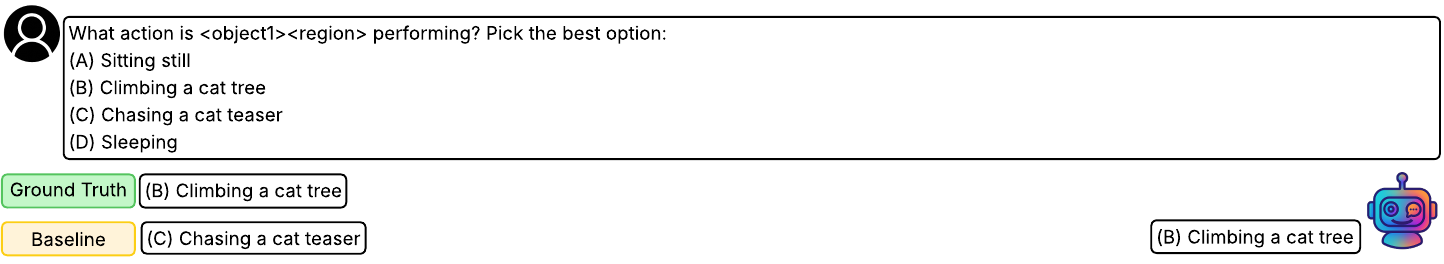}
\captionof{figure}{\textbf{Qualitative Results of Video LLMs Trained w/ and w/o \ourseos-Synthethized Data}. 
This sample is drawn from VideoRefer-Bench\textsuperscript{Q}, designed to assess a model's performance on the task of \textbf{Mask-Referred Regional QA}. 
The boundary of the region referred to by the mask in this sample is highlighted in purple.
The model trained on \ourseos-generated data correctly identifies the masklet-referred region and action.
}
    \label{fig:quali_good_benchq_cat}
\end{figure*}

\begin{figure*}[t]
	\centering
    \includegraphics[width=1.0\linewidth]{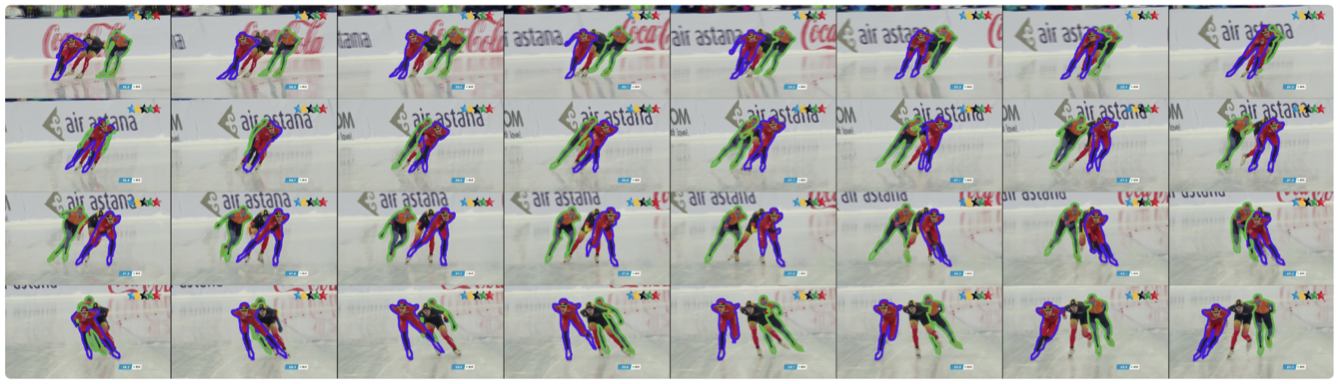}
    \includegraphics[scale=0.70]{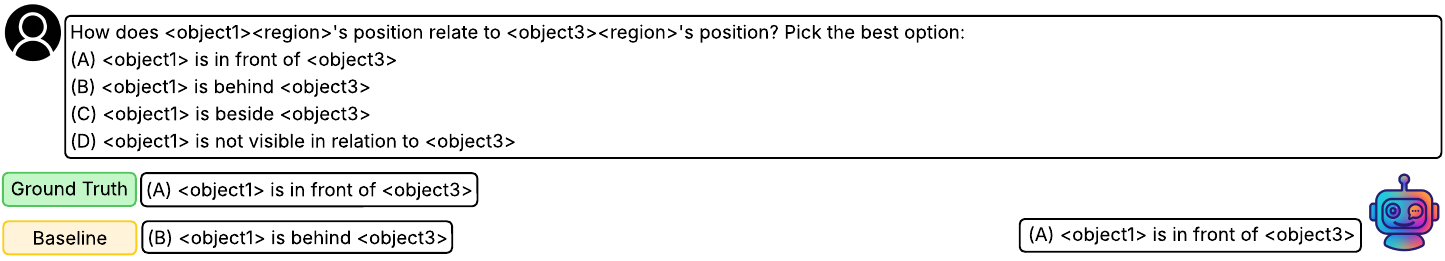}
\captionof{figure}{\textbf{Qualitative Results of Video LLMs Trained w/ and w/o \ourseos-Synthethized Data}. 
This sample is drawn from VideoRefer-Bench\textsuperscript{Q}, designed to assess a model's performance on the task of \textbf{Mask-Referred Regional QA}. 
This sample presents a multi-masklet scenario, with two masklets referring to two different individuals. 
The boundary of the $<$object1$>$ region is highlighted in purple, and $<$object2$>$ is highlighted in green.
The model trained on \ourseos-generated data correctly answers this multi-masklet reference question by effectively analyzing the relationship between the two masklets within the video context.
}
    \label{fig:quali_good_benchq_twoplayers}
\end{figure*}

\begin{figure*}[t]
	\centering
    \includegraphics[width=1.0\linewidth]{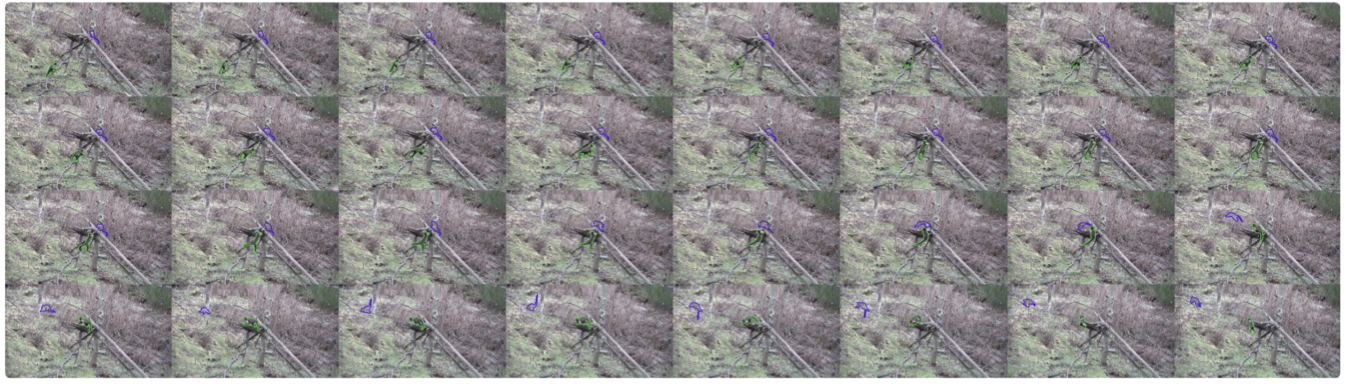}
    \includegraphics[scale=0.70]{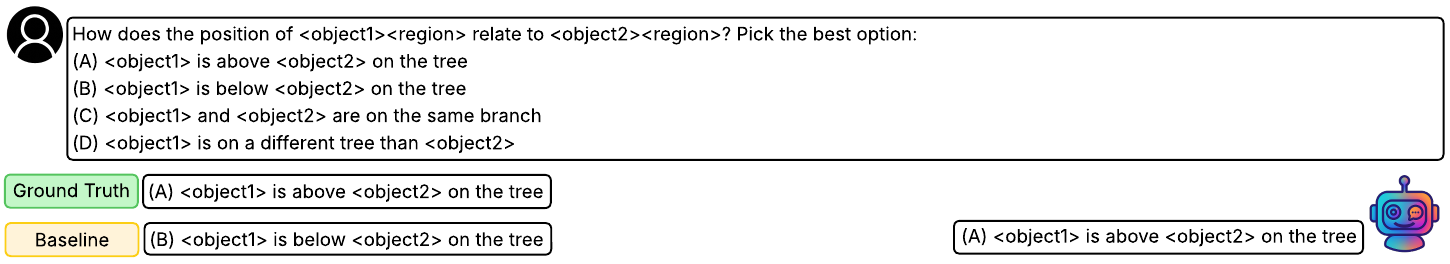}
\captionof{figure}{\textbf{Qualitative Results of Video LLMs Trained w/ and w/o \ourseos-Synthethized Data}. 
This sample is drawn from VideoRefer-Bench\textsuperscript{Q}, designed to assess a model's performance on the task of \textbf{Mask-Referred Regional QA}. 
This sample presents a multi-masklet scenario, with two masklets referring to two different individuals. 
The boundary of the $<$object1$>$ region is highlighted in purple, and $<$object2$>$ is highlighted in green. 
Kindly zoom in, as the regions are relatively small and may be difficult to discern.
The model trained on \ourseos-generated data correctly answers this multi-masklet reference question by effectively analyzing the relationship between the two masklets within the video context.
}
    \label{fig:quali_good_benchq_twomonkeys}
\end{figure*}

\begin{figure*}[t]
	\centering
    \includegraphics[width=1.0\linewidth]{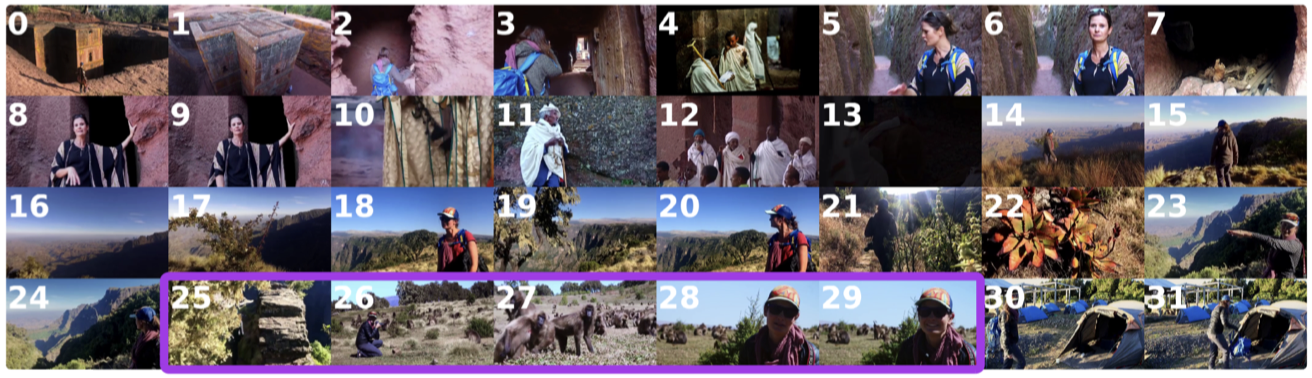}
    \includegraphics[scale=0.70]{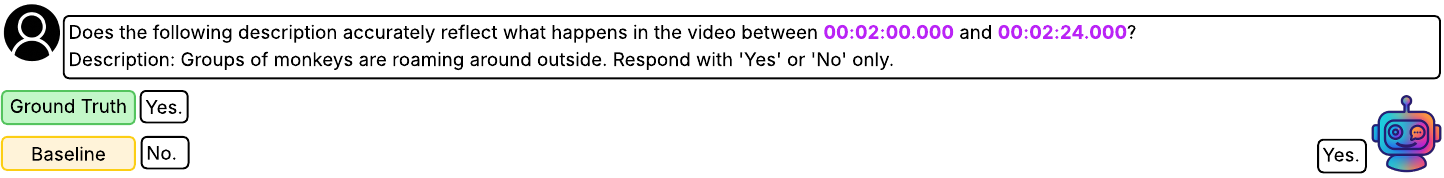}
\captionof{figure}{\textbf{Qualitative Results of Video LLMs Trained w/ and w/o \ourseos-Synthethized Data}. 
This sample is drawn from QVHighlights, using our repurposed task designed to assess a model's performance on \textbf{Timestamp-Referred Video QA}. 
The boundary of segment corresponding to the timestamps in the question is highlighted in purple. 
The model trained on our \ourseos-generated data correctly answers the question, demonstrating superior understanding of precise moments and segments in videos compared to the baseline. 
}
    \label{fig:quali_good_qv_2}
\end{figure*}

\begin{figure*}[t]
	\centering
    \includegraphics[width=1.0\linewidth]{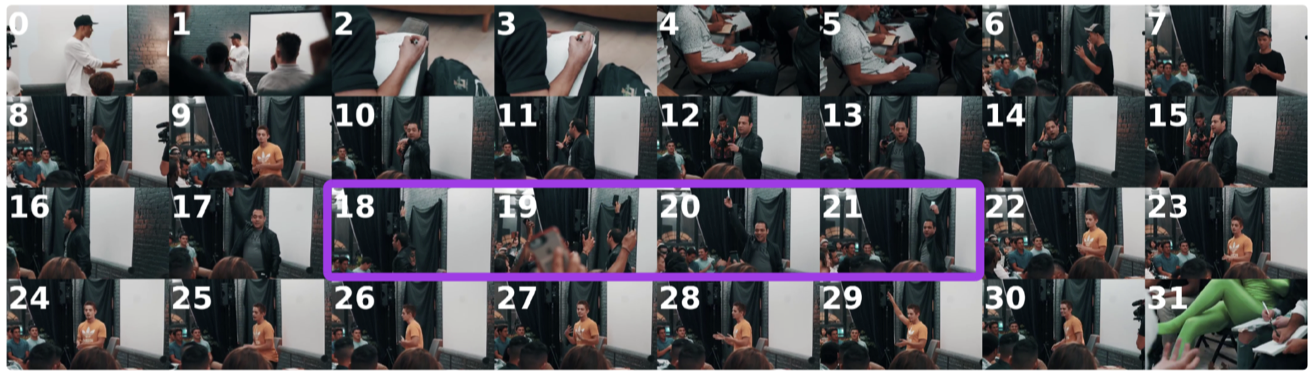}
    \includegraphics[scale=0.70]{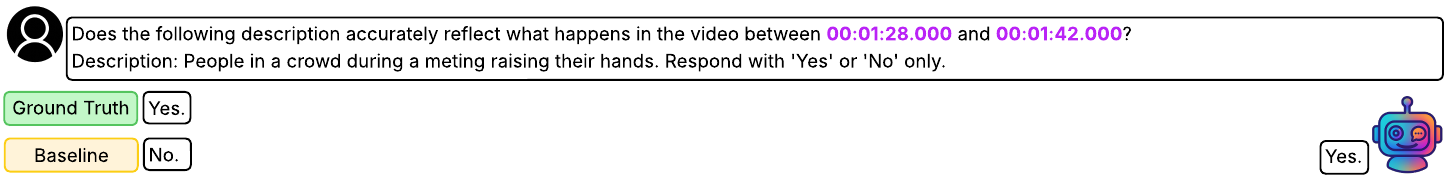}
\captionof{figure}{\textbf{Qualitative Results of Video LLMs Trained w/ and w/o \ourseos-Synthethized Data}. 
This sample is drawn from QVHighlights, using our repurposed task designed to assess a model's performance on \textbf{Timestamp-Referred Video QA}. 
The boundary of segment corresponding to the timestamps in the question is highlighted in purple. 
The model trained on our \ourseos-generated data correctly answers the question, demonstrating superior understanding of precise moments and segments in videos compared to the baseline. 
}
    \label{fig:quali_good_qv_3}
\end{figure*}

\begin{figure*}[t]
	\centering
    \includegraphics[width=1.0\linewidth]{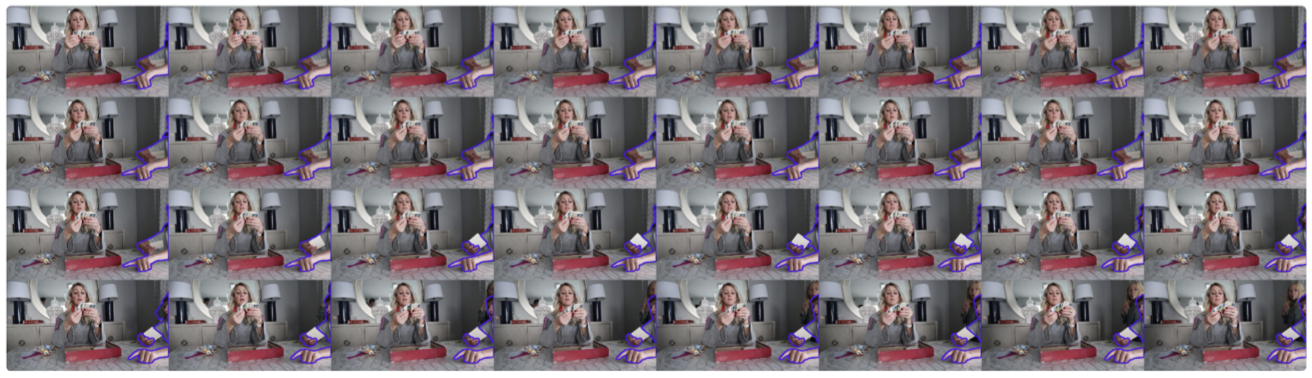}
    \includegraphics[scale=0.70]{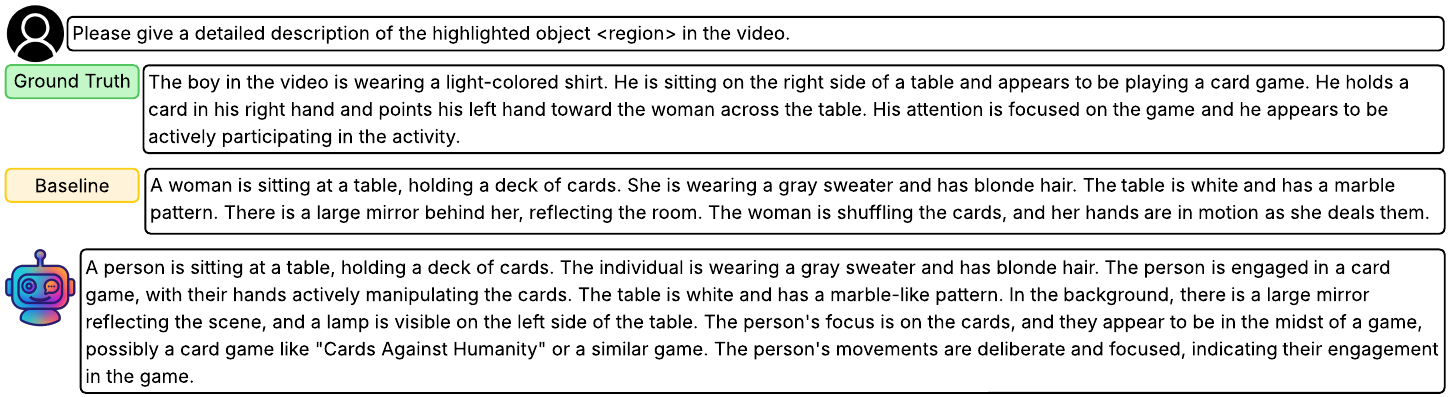}
\captionof{figure}{\textbf{Failure Results of Video LLMs Trained w/ and w/o \ourseos-Synthethized Data}. 
This sample is drawn from VideoRefer-Bench\textsuperscript{D}, designed to assess a model's performance on the task of \textbf{Mask-Referred Regional Description}. 
The boundary of the region referred to by the mask in this sample is highlighted in purple. 
The masklet is intended to refer to the boy on the right, but he is mostly out of view, while a woman appears prominently in the center of the video. Both the baseline model and the model trained on \ourseos-generated data fail to correctly interpret the masklet.
}
    \label{fig:quali_failure_1}
\end{figure*}

\begin{figure*}[t]
	\centering
    \includegraphics[width=1.0\linewidth]{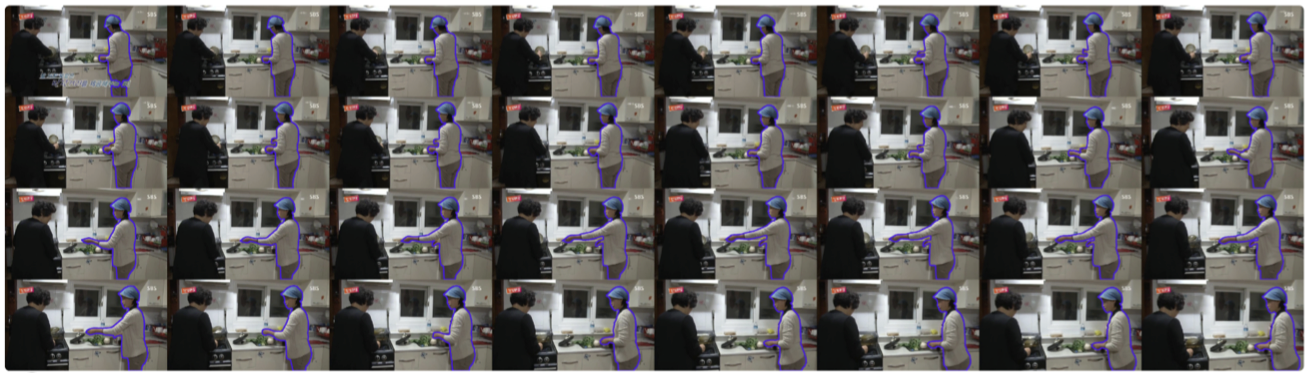}
    \includegraphics[scale=0.70]{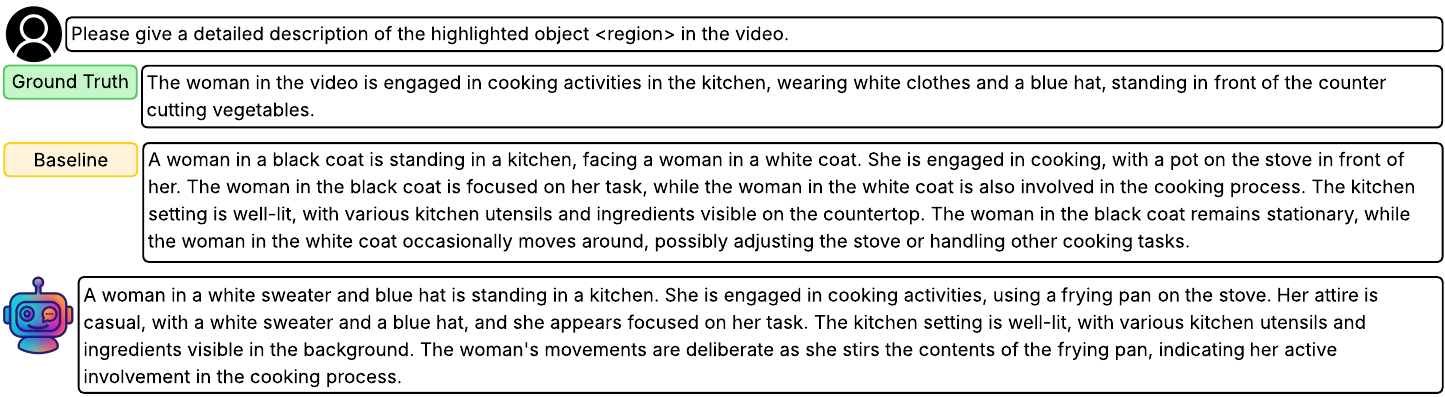}
\captionof{figure}{\textbf{Failure Results of Video LLMs Trained w/ and w/o \ourseos-Synthethized Data}. 
This sample is drawn from VideoRefer-Bench\textsuperscript{D}, designed to assess a model's performance on the task of \textbf{Mask-Referred Regional Description}. 
The boundary of the region referred to by the mask in this sample is highlighted in purple.
While the model trained on \ourseos-generated data correctly identifies that the masklet refers specifically to the woman in the white sweater, it incorrectly responds that her action is ``frying pan''. 
}
    \label{fig:quali_failure_2}
\end{figure*}

\begin{figure*}[t]
	\centering
    \includegraphics[width=1.0\linewidth]{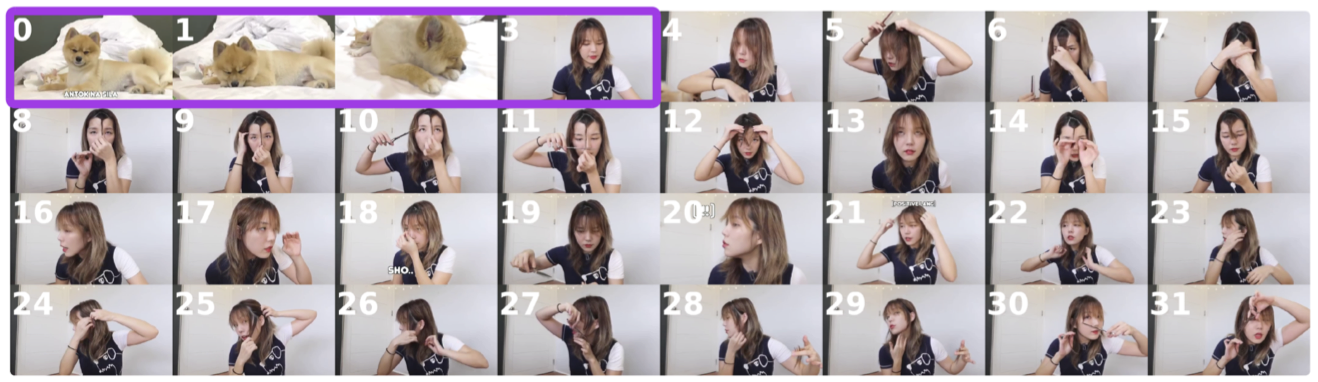}
    \includegraphics[scale=0.70]{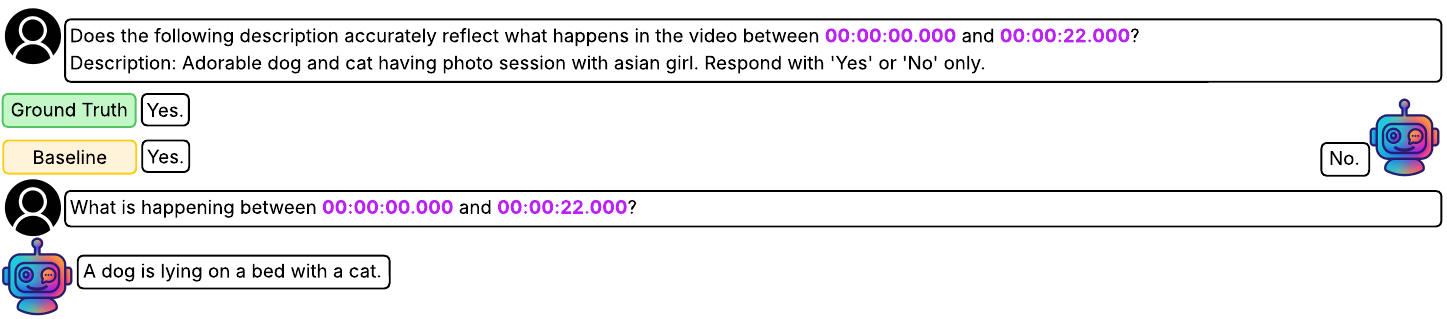}
\captionof{figure}{\textbf{Failure Results of Video LLMs Trained w/ and w/o \ourseos-Synthethized Data}. 
This sample is drawn from QVHighlights, using our repurposed task designed to assess a model's performance on \textbf{Timestamp-Referred Video QA}. 
The boundary of segment corresponding to the timestamps in the question is highlighted in purple. 
Although the model trained on our \ourseos-generated data fails to answer the question correctly, it does accurately recognize that the segment shows a dog lying on a bed with a cat. We suspect the model's failure stems from its disagreement with the description, which is not fully grounded in the visual content—for example, the video segment does not clearly depict a photo session involving the dog, the cat, and the girl. 
}
    \label{fig:quali_failure_qv_1}
\end{figure*}

We present qualitative results of our final \ourseos-trained model in comparison to the `Baseline' model (from our quantitative result tables). Specifically, results are shown in Fig.~\ref{fig:quali_good_benchd_table} and Fig.~\ref{fig:quali_good_benchd_parking} for the task of \textbf{Video Regional Captioning/Description}; Fig.~\ref{fig:quali_good_benchq_panda}, Fig.~\ref{fig:quali_good_benchq_bike}, Fig.~\ref{fig:quali_good_benchq_cat}, Fig.~\ref{fig:quali_good_benchq_twoplayers} and Fig.~\ref{fig:quali_good_benchq_twomonkeys} for the task of \textbf{Video Regional QA}; and Fig.~\ref{fig:quali_good_qv_1}, Fig.~\ref{fig:quali_good_qv_2} and Fig.~\ref{fig:quali_good_qv_3} for the task of \textbf{Timestamp-Referred Video QA}.
Three failure cases are also shown in Fig.~\ref{fig:quali_failure_1}, Fig.~\ref{fig:quali_failure_2} and Fig.~\ref{fig:quali_failure_qv_1}. Observations are in the figure captions and in Sec.~\ref{subsec:quali_results} and Sec.~\ref{sec:limit_future} of the main paper.

\begin{table*}[t]
\centering
\begin{tabular}{c}
\toprule
Prompt = ``man in the grey shirt.man in the black jacket.'' \\
\midrule
\textbf{Video Frames} \\
\includegraphics[width=0.95\linewidth]{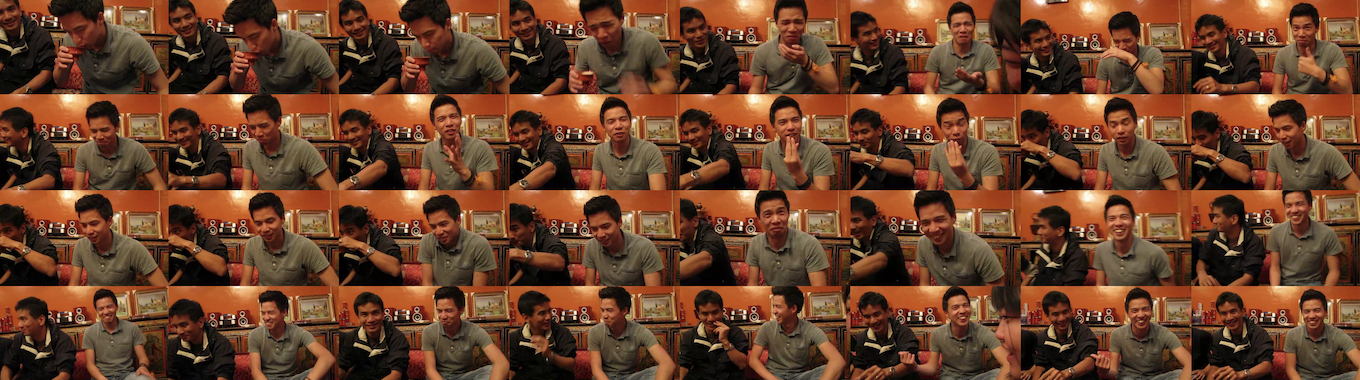}  \\
\midrule
\textbf{Ours} \\
\includegraphics[width=0.95\linewidth]{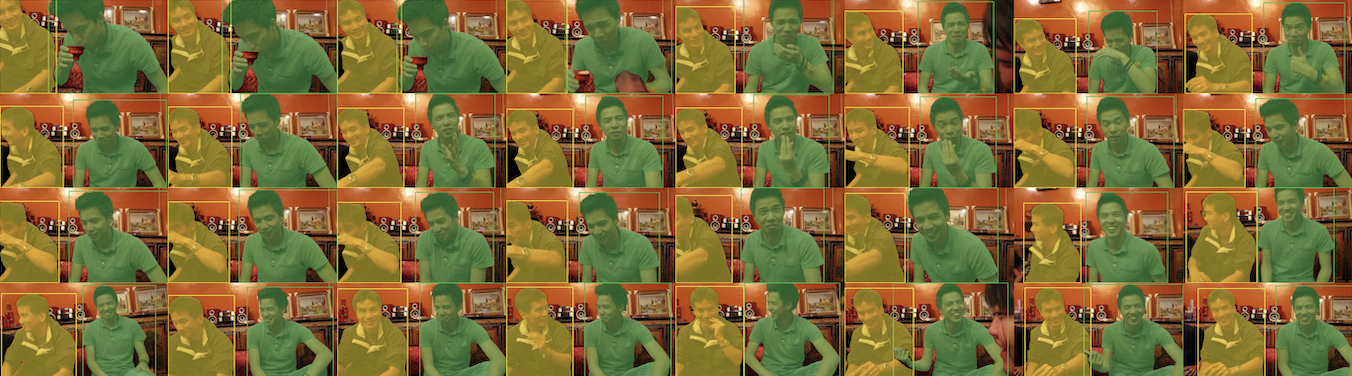}  \\
\colorbox{highlight_g}{man in the grey shirt}~\colorbox{highlight_y}{man in the black jacket} \\
\midrule
\textbf{Prior Method - GroundedSAM2~\cite{ren2025grounded_sam2}} \\
\includegraphics[width=0.95\linewidth]{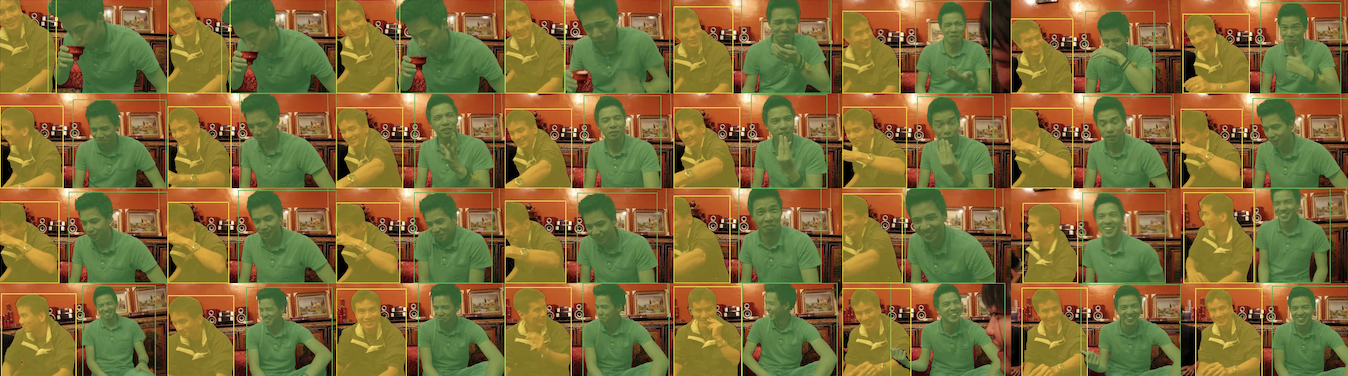}  \\
\colorbox{highlight_g}{man}~\colorbox{highlight_y}{man} \\
\bottomrule
\end{tabular}
\captionof{figure}{\textbf{Qualitative Results of Referring Masklet Generation.} In this video, our method accurately generates masklets corresponding to the input referring expressions. In contrast, GroundedSAM2~\cite{ren2025grounded_sam2} fails to differentiate between the man in the grey shirt and the man in the black jacket.}
\label{tab:quali_r2m_1}
\end{table*}

\begin{table*}[t]
\centering
\vspace{-30pt}
\begin{tabular}{c}
\toprule
Prompt = ``bride.groom.bridesmaid.officiant.'' \\
\midrule
\textbf{Video Frames} \\
\includegraphics[width=0.89\linewidth]{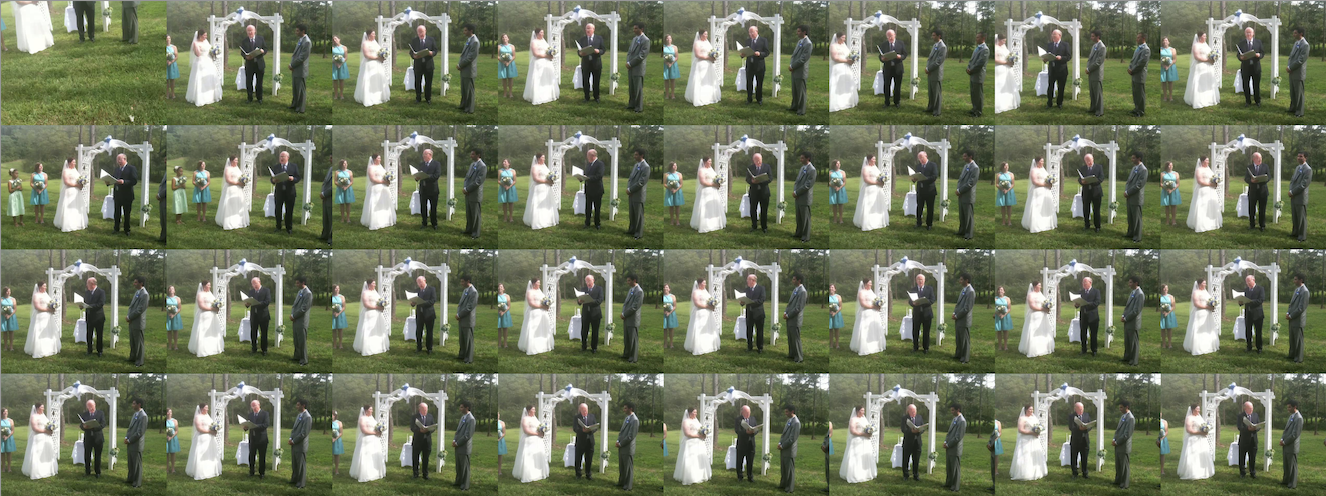}  \\
\midrule
\textbf{Ours} \\
\includegraphics[width=0.89\linewidth]{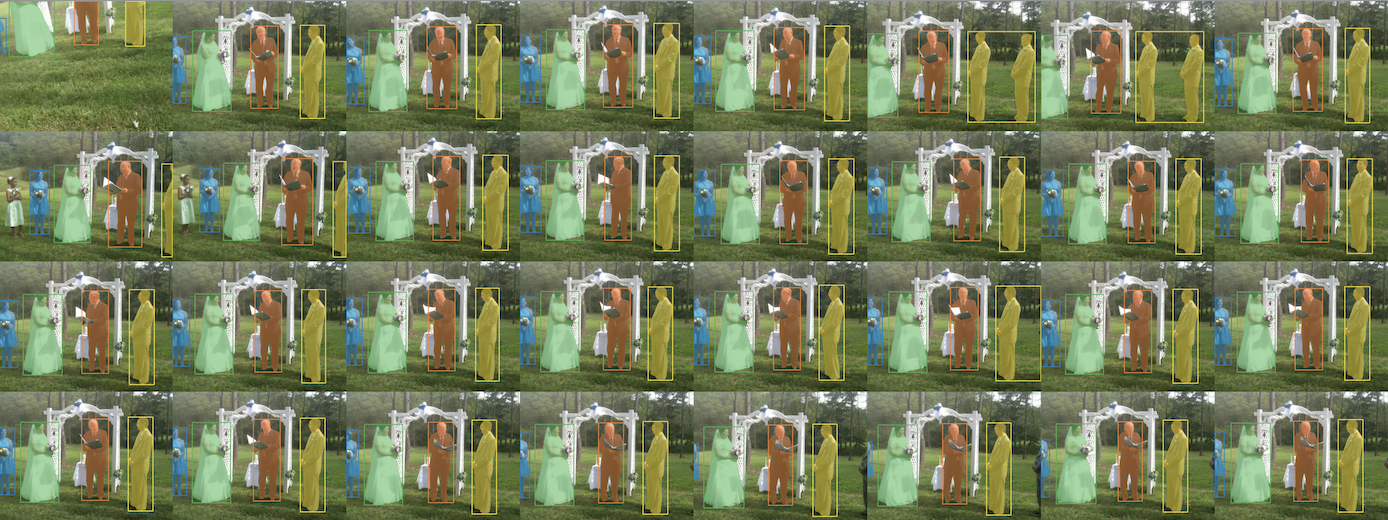}  \\
\colorbox{highlight_g}{bride}~\colorbox{highlight_y}{groom}~\colorbox{highlight_o}{officiant}~\colorbox{highlight_b}{bridesmaid} \\
\midrule
\textbf{Prior Method - GroundedSAM2~\cite{ren2025grounded_sam2}} \\
\includegraphics[width=0.89\linewidth]{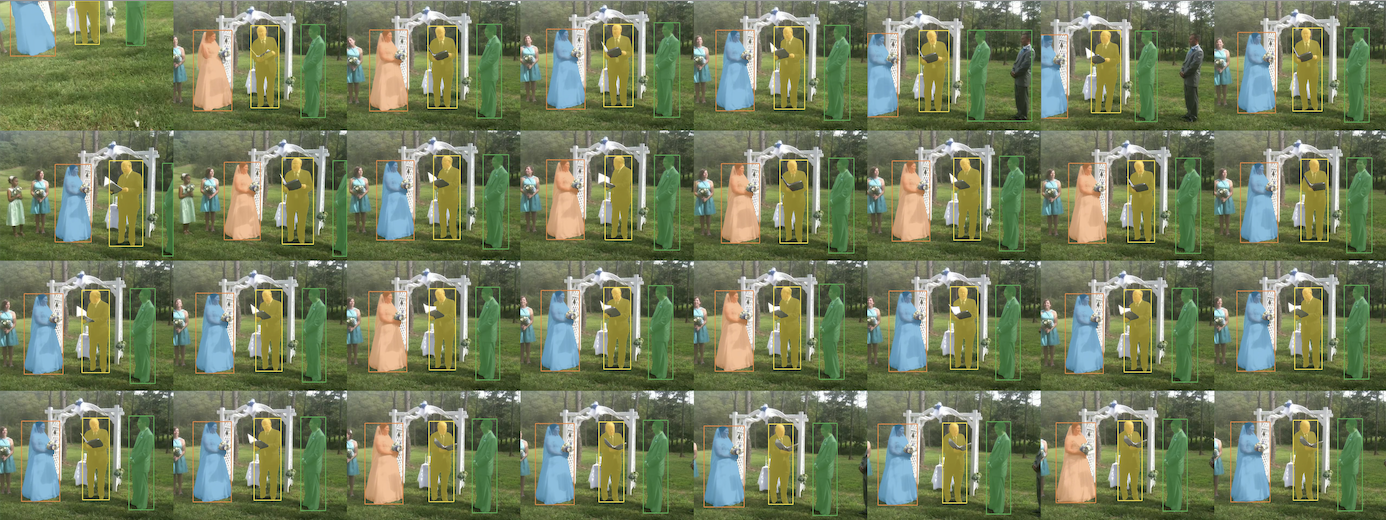}  \\
\colorbox{highlight_g}{groom officiant}~\colorbox{highlight_y}{groom}~\colorbox{highlight_o}{bride}~\colorbox{highlight_b}{bridesmaid} \\
\bottomrule
\end{tabular}
\captionof{figure}{\textbf{Qualitative Results of Referring Masklet Generation.} Our method accurately generates masklets corresponding to the input referring expressions. In contrast, GroundedSAM2~\cite{ren2025grounded_sam2} converts the input text prompt into the class names \textit{groom offician}t, \textit{groom}, \textit{bride}, and \textit{bridesmaid}. GroundedSAM2 then fails to detect the bridesmaid in the video, while incorrectly assigning the class name \textit{bridesmaid} to the actual bride, \textit{groom} to the officiant, and \textit{groom officiant} to the groom. }
\label{tab:quali_r2m_3}
\end{table*}

\subsection{Additional \ours Details}
\label{app_sec:strefer_details}

Notably, in the design of \ourseos, we choose masks to accommodate diverse, free-form spatial references from users (e.g., points, scribbles, etc.), which can be readily converted into masks using off-the-shelf tools like SAM2~\cite{sam2}.

We present our designed Referring Masklet Generation Pipeline in Algorithm~\ref{alg:referring_to_masklets} and the Video Clipper in Algorithm~\ref{alg:video_clipping}. On the right, we list the prompt used for the Video LLM-based Active Entity Recognizer.

\begin{algorithm}[t!]
\caption{Referring Masklet Generation Pipeline}
\label{alg:referring_to_masklets}
\scriptsize
\begin{algorithmic}[1]
\Require Video $\mathcal{V}$, Referring Expressions $\mathcal{R} = [r_1, \dots, r_n]$, Generalized Nouns $\mathcal{G} = [g_1, \dots, g_n]$
\Ensure Masklets aligned to each referring expression $r_i \in \mathcal{R}$

\Procedure{GenerateMasklets}{$\mathcal{V}, \mathcal{R}, \mathcal{G}$}
    \State $\mathcal{F}, S \gets$ \Call{SampleAndReorderFrames}{$\mathcal{V}$}
    \State $f^*, D_{f^*} \gets$ \Call{SelectInitialFrame}{$S, \mathcal{G}, \mathcal{R}$}
    \State $\mathcal{M} \gets$ \Call{BidirectionalSegmentationTracking}{$\mathcal{F}, f^*, D_{f^*}$}
    \State $\mathcal{M} \gets$ \Call{AssignExpressionsToMasklets}{$\mathcal{M}, f^*, \mathcal{R}, \mathcal{G}$}
    \State \Return $\mathcal{M}$
\EndProcedure

\vspace{0.5em}
\Procedure{SampleAndReorderFrames}{$\mathcal{V}$}
    \State $\mathcal{F} \gets$ sample frames from $\mathcal{V}$
    \State $S \gets$ reorder $\mathcal{F}$ using a middle-first recursive strategy
    \State \Return $(\mathcal{F}, S)$
\EndProcedure

\vspace{0.5em}
\Procedure{SelectInitialFrame}{$S, \mathcal{G}, \mathcal{R}$}
    \State max\_count $\gets -1$, $f^* \gets S[-1]$, best\_frame $\gets S[0]$, best\_detections $\gets \emptyset$
    \For{each frame $f \in S$}
        \State $D_f \gets$ \Call{GroundingDINO}{$f, \texttt{set}(\mathcal{G})$}
        \If{$|D_f| >$ max\_count}
            \State max\_count $\gets |D_f|$, best\_frame $\gets f$, best\_detections $\gets D_f$
        \EndIf
        \If{$|D_f| \geq |\mathcal{R}|$}
            \State \Return $(f, D_f)$
        \EndIf
    \EndFor
    \If{$f^* == S[-1]$ \textbf{and} best\_frame $\neq f^*$}
        \State $f^* \gets$ best\_frame, $D_{f^*} \gets$ best\_detections
    \EndIf
    \State \Return $(f^*, D_{f^*})$
\EndProcedure

\vspace{0.5em}
\Procedure{BidirectionalSegmentationTracking}{$\mathcal{F}, f^*, D_{f^*}$}
    \State Define forward sequence: $\mathcal{F}^\rightarrow = [f^*, f^*{+}1, \dots]$
    \State Define backward sequence: $\mathcal{F}^\leftarrow = [f^*, f^*{-}1, \dots]$
    \State Initialize $\mathcal{T} \gets []$
    \For{each clip $\mathcal{C} \in \{\mathcal{F}^\rightarrow, \mathcal{F}^\leftarrow\}$}
        \State $\mathcal{M}_{\mathcal{C}} \gets$ \Call{SAM2}{$\mathcal{C}, D_{f^*}, \texttt{video}$}
        \State Append $\mathcal{M}_{\mathcal{C}}$ to $\mathcal{T}$
    \EndFor
    \State \Return \Call{merge\_tracking\_results}{$\mathcal{T}$}
\EndProcedure

\vspace{0.5em}
\Procedure{AssignExpressionsToMasklets}{$\mathcal{M}, f^*, \mathcal{R}, \mathcal{G}$}
    \State Partition $\mathcal{M}$ into groups based on $\mathcal{G}$
    \For{each group $G$ in partitioned $\mathcal{M}$}
        \State $\mathcal{B}_G \gets$ available bounding boxes on frame $f^*$ in group $G$
        \For{each referring expression $r_i \in \mathcal{R}$}
            \State $b_i \gets$ \Call{RexSeek}{$f^*, \mathcal{B}_G, r_i$}
            \State Assign $r_i$ to mask in $G$ corresponding to box $b_i$
        \EndFor
        \State Post-process assignments to ensure valid mapping
    \EndFor
    \State \Return $\mathcal{M}$
\EndProcedure
\end{algorithmic}
\end{algorithm}

\begin{algorithm}[ht]
\caption{Video Clipping Pipeline}
\begin{algorithmic}[1]
\scriptsize
\Procedure{ClipVideo}{$video$}
    \State $B \gets$ \Call{PySceneDetect}{$video$, threshold=20}
    \If{$B = \emptyset$ \textbf{and} \Call{Duration}{$video$} $\geq 3$ sec}
        \State $E \gets$ \Call{GetEmbeddings}{$video$}; 
        \State $D \gets$ \Call{PairwiseDistances}{$E$}
        \State $T \gets$ \Call{ClusteringAutoThreshold}{$D$, 1.7};
        \State $C \gets$ \Call{HierarchicalAgglomerativeClustering}{$E$, $T$}
        \State $B \gets$ \Call{ExtractClipTimestampBoundaries}{$C$}
    \EndIf
    \State \Return $B$
\EndProcedure

\vspace{0.5em}
\Procedure{ClusteringAutoThreshold}{$D$, $f$}
    \State $m \gets$ mean($D$); \quad $s \gets$ std($D$); \quad $M \gets$ max($D$)
    \State \Return $\min(m + f \cdot s,\ M)$
\EndProcedure
\end{algorithmic}
\label{alg:video_clipping}
\end{algorithm}

\begin{tcolorbox}[
  colback=lightgray!20,
  colframe=black,
  arc=4mm,
  boxrule=0.8pt,
  width=\columnwidth,
  enlarge left by=0mm,
  enlarge right by=0mm,
  title=Prompt P.1: Entity Recognizer,   
  label={prompt_box:entity_recognizer}   
]
Prompt: What ``active'' scene entities can you identify from the video? An entity refers to an object, and ``active'' scene entities are scene objects that have any dynamic behaviors, such as actions, interactions with others, or movements. Please compile a list of clearly visible ``active'' scene entities from the video. Use entity appearance in concise description to distinguish one ``active'' scene entity from another if possible.
\end{tcolorbox}

\subsection{Model Details}
\label{app_sec:model_details}

\subsubsection{Architecture Overview}
\label{app_subsec:arch_overview}
The Video LLM processes a video and a user's multimodal query to generate a textual response. A multimodal query consists of the textual component of the question, a masklet along with its associated frames referring to a specific region within the video, and optionally, one or more specific timestamps within the video.

The architecture of the Video LLM is illustrated in Fig.~\ref{fig:model_architecture}. 
At a high level, the LLM processes four types of input tokens: 
(i) \emph{visual tokens}, which encode the global context of the video; 
(ii) \emph{region tokens}, which represent specific visual regions referenced in the user query (e.g., a mask or masklet); 
(iii) \emph{timestamp/temporal tokens}, which indicate particular temporal locations within the video; and 
(iv) \emph{text tokens}, which represent the textual content of the query itself. 
These tokens are jointly fed into the LLM, which then auto-regressively generates a textual response. 

The construction of visual, region, and timestamp tokens from raw inputs—namely, the video and multimodal user query—is detailed in Sec.~\ref{app_subsec:video_token}, Sec.~\ref{app_subsec:mask_token}, and Sec.~\ref{app_subsec:timestamp_token}, respectively.

\subsubsection{Video Token Representation}
\label{app_subsec:video_token}
Given an input video $x_{\mathrm{v}} \in \mathbb{R}^{T_{v} \times 3 \times H_{v} \times W_{v}}$, where $T_v$ is the number of frames and $H_v,W_v$ are the height and width of the frames, a visual encoder extracts the video's global visual features $f_{\mathrm{v}} \in \mathbb{R}^{t_{v} \times d_{v} \times h_{v} \times w_{v}}$.

A Video-Language Connector is then applied on top of the visual encoder to project the global visual features into a sequence of visual tokens $e_{\mathrm{v}} \in \mathbb{R}^{L_{v} \times d}$, where $d$ represents the dimensionality of the language model's input token space, and $L_{v}$ is the number of visual tokens of a video. This connector aligns the visual features to the input space of a language model while preserving semantics relevant for multimodal understanding. In some designs (e.g., BLIP-3-Video~\cite{blip3video} that our model architecture is based on), the connector also incorporates a token compression module to reduce the number of tokens, improving efficiency without sacrificing critical information.

\subsubsection{Masklet Reference Token Representation}
\label{app_subsec:mask_token}
Our modified Video LLM is designed to understand user queries about videos that involve spatial or spatiotemporal, local regional references. To support diverse, free-form spatial reference from users (e.g., points, scribbles, etc.), we standardize them by converting these free-form spatial references into regional masks before processed by the model. This approach is effective because many forms of spatial reference can be easily transformed into masks using off-the-shelf tools like SAM2~\cite{sam2}. 

\vspace{3pt}
\noindent \textbf{Mask and Masklet}. A regional mask is represented as a 2D binary matrix $~\mathbb{R}^{H_{m} \times W_{m}}$, where $H_m$ and $W_m$ are the height and width of the image containing the region of interest, with a value of $1$ inside the region and $0$ outside. When extended over time, a temporal sequence of such regional masks  $x_{\mathrm{r}} \in \mathbb{R}^{T_{m} \times H_{m} \times W_{m}}$
is referred to as a masklet. Since a mask is special case of masklet with only one frame, we describe the masklet feature extraction process below.

\vspace{3pt}
\noindent \textbf{Masklet Token Representation}. Leveraging the same visual encoder, our model extracts image feature maps $f_{\mathrm{m}} \in \mathbb{R}^{t_{m} \times d_{v} \times h_{m} \times w_{m}}$ for the frames that contain the masklet $x_{\mathrm{r}}$. The masklet $x_{\mathrm{r}}$ and its corresponding frames' feature maps $f_{\mathrm{m}}$ are then processed by a Region-Language Connector, which outputs region tokens $e_{\mathrm{r}} \in \mathbb{R}^{L_{r} \times d}$ that are aligned to the language space, where $L_{r}$ is a predefined number of region tokens.

The Region-Language Connector begins by resizing the binary masklet $x_{\mathrm{r}}$ via bilinear interpolation to match the 
spatial (and temporal if the visual encoder condenses the time axis) dimensions of $f_{\mathrm{m}}$, yielding a resized masklet of shape $\mathbb{R}^{t_{m} \times h_{m} \times w_{m}}$. A Mask Pooling operation is then applied: average pooling is performed over the spatial locations within the mask region for each frame, producing a pooled feature representation $p \in \mathbb{R}^{t_{m} \times d_{v}}$. This representation can be interpreted as a sequence of $t_{m}$ region tokens, each of dimensionality $d_{v}$.

To reduce the temporal redundancy, a Temporal Token Merge module condenses the $t_{m}$ tokens into $L_{r}$ representative ones ($L_{r}<t_{m}$). Specifically, for $p \in \mathbb{R}^{t_m \times d_v}$, cosine similarities are computed between each pair of temporally adjacent tokens:

\begin{equation}
    \mathbf{s}_{i, i+1} = \frac{p^i \cdot p^{i+1}}{\left\|p^i\right\| \cdot \left\|p^{i+1}\right\|}, \quad 0 \leq i < t_m - 1
\end{equation}

This yields a similarity vector $\mathbf{s} \in \mathbb{R}^{t_m - 1}$.  
A similarity threshold $\theta$ is then selected as the $L_r$-th largest value in $\mathbf{s}$.  
Next, the sequence $p$ is processed sequentially from the beginning to the end to form token groups. An initially empty group is created and the first token in $p$ is added to it. For each index $i$ from $0$ to $t_m - 2$, if $\mathbf{s}_{i, i+1} \geq \theta$, then $p^{i+1}$ is added to the current group. Otherwise, the current group is finalized, and a new group is initiated with $p^{i+1}$. 

This process produces exactly $L_{r}$ token groups.  
Each group is finally merged into a single representative token by averaging the embeddings of all tokens within the group.

Finally, the resulting $L_{r}$ tokens, each in $\mathbb{R}^{d_{v}}$, are projected into the language embedding space via an MLP, producing the final region tokens $e_{\mathrm{r}} \in \mathbb{R}^{L_{r} \times d}$.

Note that the Temporal Token Merge module is bypassed when the user query involves only a single frame mask (as opposed to a masklet).

\subsubsection{Timestamp Reference Token Representation}
\label{app_subsec:timestamp_token}
By design, our model is effective in scenarios where  
users may refer to specific times within videos in their queries. However, LLMs often struggle with interpreting numerical values~\cite{schwartz2024numerologic}. To address this challenge, we adopt the Temporal Token Representation method introduced in Grounded-VideoLLM~\cite{wang2024grounded}, which discretizes continuous time into a sequence of temporal tokens, making time-related reasoning more manageable for LLMs.

Suppose the video has a duration of $L$ seconds. We divide it into $M$ equal-length, non-overlapping, and non-spacing segments, resulting in $M+1$ anchor points that span from the start to the end of the video. These anchor points, labeled from \textless$0$\textgreater~to \textless$M$\textgreater, represent evenly spaced temporal positions throughout the video. 
Each specific timestamp within the video is mapped to an anchor point and then encoded as a temporal token. For example, \textless$0$\textgreater~marks the beginning of the video, while \textless$M$\textgreater~represents the end. 
These $M+1$ anchor points are added to the LLM's vocabulary (by expanding the LLM's vocabulary), 
enabling unified modeling of time alongside text. Mathematically:

A specific continuous timestamp $\tau$ can be easily converted to a temporal token \textless$t$\textgreater{} and vice versa:
\begin{equation}
    t=\operatorname{Round}\left(M \cdot \frac{\tau}{L}\right), \quad \tau=L \cdot \frac{t}{M}
\end{equation}
In this way, specific timestamps in the user query are converted into timestamp anchor points. Both text and timestamp anchor points are then mapped to embeddings through the extended word embedding layer of the LLM, forming interleaved text tokens and temporal tokens.

In our model tuning and evaluation experiments, since our Video LLM processes $32$ input frames, we set $M=31$ to learn $32$ temporal tokens.

\clearpage
\begin{table*}[!t]
\centering
\renewcommand{\arraystretch}{1.2}
\scriptsize
\vspace{-20pt}
\begin{tabular}{|p{3cm}|p{1.5cm}|p{0.8cm}|p{0.8cm}|p{4.5cm}|p{4.5cm}|}
\hline
\textbf{Type ID \& Task} & \textbf{Frames} & \textbf{Source} & \textbf{Format} & \textbf{Example Question-Answer Pair} & \textbf{Mask-Refer Version} \\
\hline
 1. Ask the model to describe the behavior of entities that are present in a segment of the video. & Frames are extracted only from the segment of the video. & Template & OE & \underline{Question}: {\textless}video{\textgreater}What is happening to the woman? \newline 
\underline{Answer}: The woman is engaged in a dance with the man, involving spins and turns. She is lifted off the ground by the man during the dance. & \underline{Question}:{\textless}video{\textgreater}Please answer the following question about the {\textless}region{\textgreater}. What is happening to her? \newline 
\underline{Answer}: The woman is engaged in a dance with the man, involving spins and turns. She is lifted off the ground by the man during the dance. \\
\hline
 2. Ask the model to describe the behavior of entities that are not present in a segment of the video; the model should respond with uncertainty (e.g., ``Sorry, I’m not sure''). & Frames are extracted only from the segment of the video. & Template & OE & \underline{Question}: {\textless}video{\textgreater}What is currently happening to the person in a green hoodie?\newline \underline{Answer}: The person in a green hoodie seems to be not clearly visible. & N/A \\
\hline
 3. Ask a yes/no question about the presence of an entity in a segment of the video; if present, the model should describe its behavior; if absent, the model should respond with uncertainty. & Frames are extracted only from the segment of the video. & Template & OE & \underline{Question}: {\textless}video{\textgreater}Were you able to see a woman in a black jacket?\newline \underline{Answer}: Yes. The woman walks towards the child seated on the sofa. & N/A \\
\hline
 4. Ask a yes/no question about the presence of an entity in a segment of the video; the model should respond with a concise “Yes” or “No” only. & Frames are extracted only from the segment of the video. & Template & OE & \underline{Question}:{\textless}video{\textgreater} Is there a woman in a black jacket? Answer only ``Yes'' or ``No''.\newline \underline{Answer}: Yes. & N/A \\
\hline
 5. Ask the model to identify the correct temporal order in which entities first appear in the video from multiple choices. & Frames are extracted from the full video. & Template & MCQ & \underline{Question}:{\textless}video{\textgreater} Which order shows their first appearance in the video?\newline
(A) child interacting with the plant bed, child holding a bag and a toy, child walking across the lawn\newline
(B) child holding a bag and a toy, child approaching a plant bed, child interacting with the plant bed\newline
(C) child approaching a plant bed, child holding a bag and a toy, child interacting with the plant bed\newline
(D) child walking across the lawn, child holding a bag and a toy, child interacting with the plant bed\newline
\underline{Answer}: (B) & N/A \\
\hline
 6. Ask the model to describe the behavior of entities that may or may not be present in a specific time range of the video; the question refers to a time range. & Frames are extracted from the full video. & Template & OE & \underline{Question}:{\textless}video{\textgreater} Could you explain what the girl in the yellow coat is doing between 00:00:05 and 00:00:12.210?\newline
\underline{Answer}: The girl in the yellow coat is carefully watering plants in a garden. & \underline{Question}:{\textless}video{\textgreater}Please answer the following question about the {\textless}region{\textgreater}. Could you explain what she is doing between 00:00:05 and 00:00:12.210?\newline
\underline{Answer}: The girl in the yellow coat is carefully watering plants in a garden. \\
\hline
 7. Ask the model to describe what happened generally or to a specific entity during a specific time range in the video; the question refers to a time range. & Frames are extracted from the full video. & LLM & OE \& MCQ & \underline{Question}:{\textless}video{\textgreater} What else did the woman interviewing the man do between 00:00:00 and 00:00:07.007?\newline
\underline{Answer}: The woman interviewing the man is talking as well. & \underline{Question}:{\textless}video{\textgreater}Please answer the following question about the {\textless}region{\textgreater}. What else did she do between 00:00:00 and 00:00:07.007?\newline
\underline{Answer}: The woman interviewing the man is talking as well.  \\
\hline
 8. Ask the model to describe what happened generally or to a specific entity during a coarse time range in the video (e.g., throughout the video, beginning, middle, or end). & Frames are extracted from the full video. & LLM & OE \& MCQ & \underline{Question}:{\textless}video{\textgreater} What else did the woman interviewing the man do in the beginning of the video?\newline \underline{Answer}: The woman interviewing the man is talking as well. & \underline{Question}:{\textless}video{\textgreater}Please answer the following question about the {\textless}region{\textgreater}. What else did she do in the beginning of the video?\newline \underline{Answer}: The woman interviewing the man is talking as well. \\
\hline
 9. Ask the model to identify when a specific behavior or event occurs within the video; expect the model to answer with a coarse time range in the video (e.g., throughout the video, beginning, middle, or end).
& Frames are extracted from the full video. & LLM & OE \& MCQ & \underline{Question}:{\textless}video{\textgreater} During which part of the video was the child in pink dress riding a tricycle?\newline
\underline{Answer}: The beginning. & \underline{Question}:{\textless}video{\textgreater}Please answer the following question about the {\textless}region{\textgreater}. During which part of the video was this person riding a tricycle?\newline
\underline{Answer}: The beginning. \\
\hline
 10. Ask the model to describe the behavior of an entity before/during/after something else occurs.  & Frames are extracted from the full video. & LLM & OE \& MCQ & \underline{Question}:{\textless}video{\textgreater} What is the adult doing while the child is riding a tricycle?\newline
\underline{Answer}: The adult is watching and walking behind the child. & \underline{Question}:{\textless}video{\textgreater}Please answer the following question about the {\textless}region{\textgreater}. What is he doing while the child is riding a tricycle?\newline
\underline{Answer}: The adult is watching and walking behind the child. - \\
\hline
 11. Ask the model to identify the entity involved before/during/after something else occurs.  & Frames are extracted from the full video. & LLM & OE \& MCQ & \underline{Question}: Who is walking behind the child in blue while the child is riding a tricycle?\newline
\underline{Answer}: An adult wearing a black shirt.  & \underline{Question}:{\textless}video{\textgreater}Please answer the following question about the {\textless}region{\textgreater}. Who is walking behind this child while the child is riding a tricycle?\newline
\underline{Answer}: An adult wearing a black shirt. \\
\hline
\end{tabular}
\vspace{-5pt}
\caption{\textbf{Details of \ourseos-synthesized video instruction data}. The table details the question task types, their visual inputs, QA generation sources, formats, examples, and the mask-referring versions of the QAs.
`OE' denotes open-ended QA, and `MCQ' indicates multiple-choice QA.}
\label{tab:question_types}
\end{table*}

\end{document}